\documentclass[10pt,twocolumn,letterpaper]{article}

\usepackage{cvpr}              

\usepackage[dvipsnames]{xcolor}

\definecolor{cvprblue}{rgb}{0.21,0.49,0.74}

\usepackage{graphicx}
\usepackage{amsmath}
\usepackage{amssymb}
\usepackage{booktabs}
\usepackage{multirow}
\usepackage{float}

\usepackage{color, colortbl}
\usepackage{pifont}
\usepackage{makecell}
\usepackage{adjustbox}
\usepackage{svg}

\usepackage[pagebackref,breaklinks,colorlinks]{hyperref}

\def\paperID{2177} 
\def\confName{CVPR}
\def\confYear{2024}

\title{ 
MP5: A Multi-modal Open-ended Embodied System \\ in Minecraft via Active Perception}

\author{
Yiran Qin\textsuperscript{1,2}\footnotemark[1], \quad
Enshen Zhou\textsuperscript{1,3}\footnotemark[1], \quad
Qichang Liu\textsuperscript{1,4}\footnotemark[1], \quad
Zhenfei Yin\textsuperscript{1,5}, \\
Lu Sheng\textsuperscript{3}\footnotemark[2], \quad
Ruimao Zhang\textsuperscript{2}\footnotemark[2], \quad
Yu Qiao\textsuperscript{1}, \quad
Jing Shao\textsuperscript{1}\footnotemark[3]\\
\small $^{1}$Shanghai Artificial Intelligence Laboratory~~
\small$^{2}$The Chinese University of Hong Kong, Shenzhen \\
\small$^{3}$School of Software, Beihang University ~~
\small$^{4}$Tsinghua University~~~
\small$^{5}$The University of Sydney\\
\tt\footnotesize yiranqin@link.cuhk.edu.cn~~~zhouenshen@buaa.edu.cn~~~\\
\tt\footnotesize
lsheng@buaa.edu.cn~~~ruimao.zhang@ieee.org~~~shaojing@pjlab.org.cn\\
}

\newcommand{\mname}{\textsl{MP5}}

\begin{document}

\twocolumn[{%
\maketitle

\begin{figure}[H]
\vspace{-8mm}
\hsize=\textwidth 
\centering
\includegraphics[width=\textwidth]{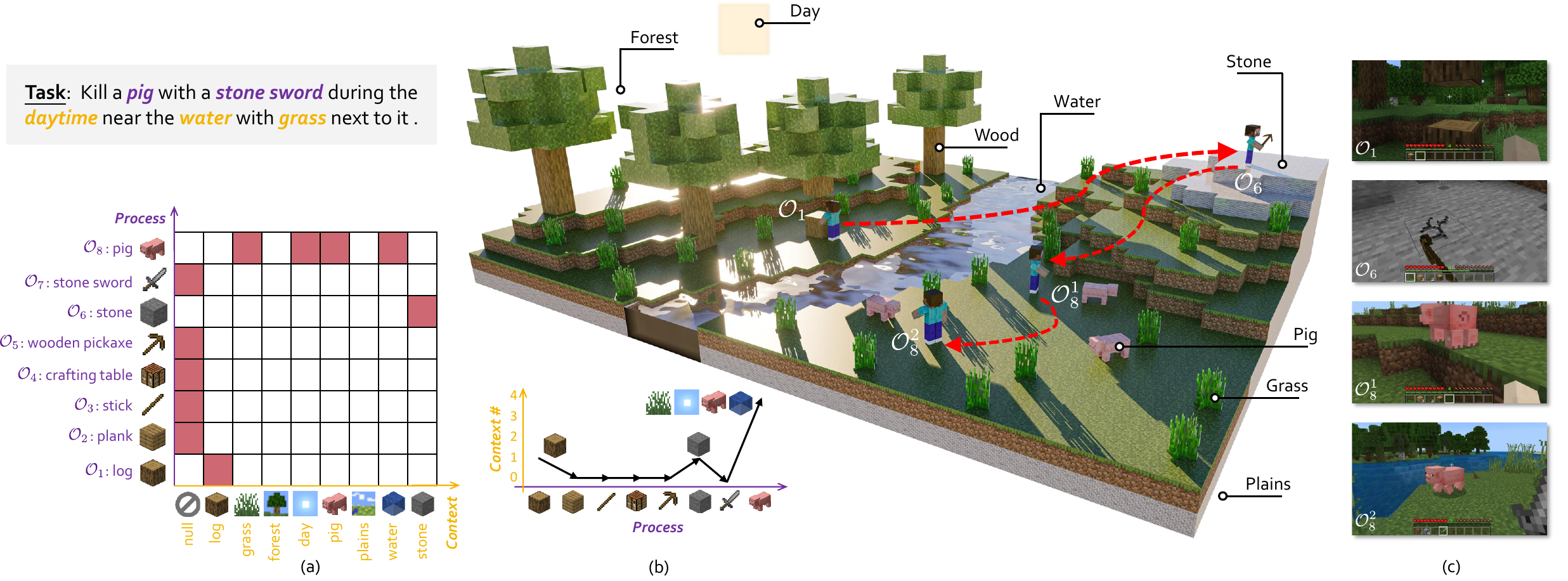}
%
\caption{The process of finishing the task \textit{``kill a pig with a stone sward during the daytime near the water with grass next to it.''}.
\textbf{(a)} To achieve the final goal (\ie, $\mathcal{O}_8$: ``kill a pig~\raisebox{-0.3ex}{\includegraphics[width=0.35cm]{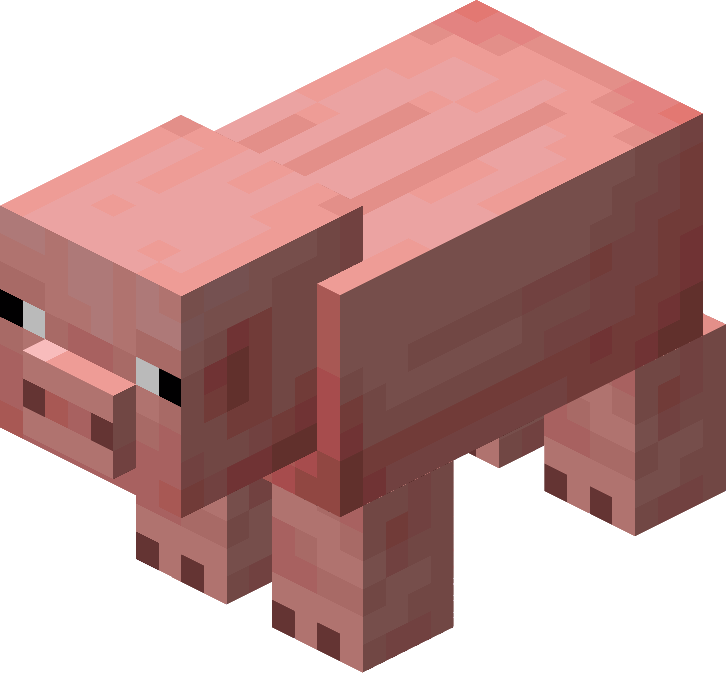}}''), a player should accomplish a list of sub-objectives $\{\mathcal{O}_i\}_{i=1}^7$ sequentially. During this process, the player should also be aware of some items in the environment, \eg, ``grass~\raisebox{-0.3ex}{\includegraphics[width=0.3cm]{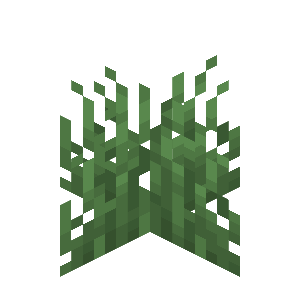}}'', ``day~\raisebox{-0.3ex}{\includegraphics[width=0.3cm]{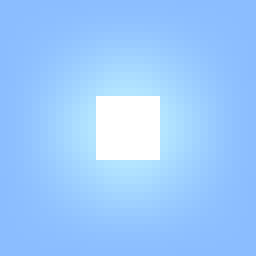}}'' and \etc.
\textbf{(b)} This diagram shows the number of these necessary items in the context that should be perceived for each sub-objective, during the task execution process.  
\textbf{(c)} Images marked by $\mathcal{O}_1$ and $\mathcal{O}_6$  show the observed ego-centric views in the process of achieving the corresponding sub-objectives.
Images marked by $\mathcal{O}_{8}^{1}$ and $\mathcal{O}_8^{2}$ indicate how the player executes the action about the last sub-objective ``kill a pig~\raisebox{-0.3ex}{\includegraphics[width=0.35cm]{icon/pig.png}}''.
%
This exemplar process tells that such long-horizon open-world embodied tasks in Minecraft should be solved both in process-dependent and context-dependent way.
%
}
\label{fig:motivation}
\end{figure}
}]

\let\thefootnote\relax\footnotetext{$^*$ Equal contribution\hspace{3pt} \hspace{5pt}$^\dagger$ Corresponding author\hspace{5pt} $^\ddagger$ Project leader
}

\begin{abstract}

It is a long-lasting goal to design an embodied system that can solve long-horizon open-world tasks in human-like ways.
%
%
However, existing approaches usually struggle with compound difficulties caused by the logic-aware decomposition and context-aware execution of these tasks. 
To this end, we introduce {\mname}, an open-ended multimodal embodied system built upon the challenging Minecraft simulator, which can decompose feasible sub-objectives, design sophisticated situation-aware plans, and perform embodied action control, with frequent communication with a goal-conditioned active perception scheme.
Specifically, {\mname} is developed on top of recent advances in Multimodal Large Language Models (MLLMs), and the system is modulated into functional modules that can be scheduled and collaborated to ultimately solve pre-defined context- and process-dependent tasks.
Extensive experiments prove that {\mname} can achieve a $22\%$ success rate on difficult process-dependent tasks and a $91\%$ success rate on tasks that heavily depend on the context. 
Moreover, {\mname} exhibits a remarkable ability to address many open-ended tasks that are entirely novel.
Please see the project page at \url{https://iranqin.github.io/MP5.github.io/}.

\end{abstract}

\section{Introduction}
\label{sec:intro}

One of the core objectives of current embodied intelligence is to construct generalist agents that can solve long-horizon open-world embodied tasks, approaching the behavior patterns of human beings~\cite{baker2022video,oh2017zero,lin2021juewu,reed2022generalist}. 
However, the \emph{process dependency} and \emph{context dependency} in these tasks, such as those in Minecraft depicted in \cref{fig:motivation}, hinder recent agents from achieving the aforementioned goal.
%
%
To be specific, the former emphasizes the inherent dependency among the sub-objectives of one task or an action sequence to fulfill one sub-objective (such as ``craft a stone sword~\raisebox{-0.3ex}{\includegraphics[width=0.3cm]{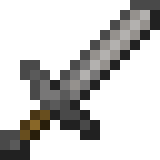}}'' should be solved before ``kill a pig~\raisebox{-0.3ex}{\includegraphics[width=0.3cm]{icon/pig.png}}''). 
The latter highlights that the execution of each sub-objective or even each action depends on the contextual information of the environment (such as ``kill a pig~\raisebox{-0.3ex}{\includegraphics[width=0.3cm]{icon/pig.png}}'' requires to find the target ``pig~\raisebox{-0.3ex}{\includegraphics[width=0.3cm]{icon/pig.png}}'' and its surrounding items ``grass~\raisebox{-0.3ex}{\includegraphics[width=0.3cm]{icon/grass.png}}'' and ``water~\raisebox{-0.3ex}{\includegraphics[width=0.3cm]{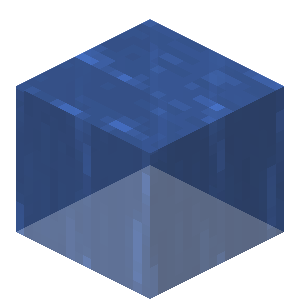}}'' during the ``daytime~\raisebox{-0.3ex}{\includegraphics[width=0.3cm]{icon/sun.png}}'' in the observed images, as shown in \cref{fig:motivation}).

%
%

The recent success of Large Language Models (LLMs) has attempted to solve the process-dependent challenge, by using LLMs to break down a long-horizon process-dependent task into a sequence of feasible sub-objectives~\cite{wang2023describe,zhu2023ghost,wang2023voyager}.
These methods~\cite{wang2023voyager,zhu2023ghost} simplify the context-dependent challenge by assuming the agents are all-seeing, \ie, knowing everything about their state and the environment it locates in.
%
%
%
%
However, to solve the context-dependent challenge, an embodied agent should additionally have: (1) the perception capability is open-ended, selective and give results tailored to diverse purposes (\eg, for task planning or action execution), (2) the perception module can be compatibly scheduled along with the other modules (\eg, planning and execution modules) by a unified interface, as an integrated system.

To this end, we introduce {\mname}, a novel embodied system developed within Minecraft, to meet the above expectations.
Specifically, {\mname} comprises five interacting modules, \ie,
\textbf{Parser} decomposes a long-horizon task into a sequence of sub-objectives that should be completed one by one;
\textbf{Percipient} answers various questions about the observed images, as the reference for the other modules;
\textbf{Planner} schedules the action sequences of a sub-objective, as well as refines the following sub-objectives, given the current situation;
\textbf{Performer} executes the actions along with frequent interaction with the environment;
and \textbf{Patroller} checks the responses from the Percipient, Planner, and Performer, for the purpose of verifying current plans/actions, or feedback on potential better strategies.
In our work, Percipient is a LoRA-enabled Multimodal LLM (MLLM). Among the pre-trained LLMs, Parser and Planner are augmented with external Memory, while Patroller is not.
%
%

%
Notably, {\mname} includes an \emph{active perception} scheme by means of multi-round interaction between {Percipient} and {Patroller}, which is to actively perceive the contextual information in the observed images, with respect to various queries raised by Planner and Performer. 
It is the key enabler to solve context-dependent tasks. 
Patroller in this scheme relays compatible feedback to Planner and Performer accordingly, while eventually strengthening the planning skill in awareness of the situations and improving the action execution correctness in an embodied manner.

Extensive experiments prove that {\mname} can robustly complete tasks needed for long-horizon reasoning and complex context understanding. It achieved a $22\%$ success rate on diamond-level tasks (\ie, one of the hardest long-horizon tasks) and a $91\%$ success rate on tasks requiring complex scene understanding (\ie, need to perceive around $4\sim 6$ key items in the observed images). 
Moreover, in \cref{sub:open_ended_tasks}, {\mname} can surprisingly address more open-end tasks both with heavy process dependency and context dependency.

\begin{figure*}
\centering
\includegraphics[width=1\linewidth]{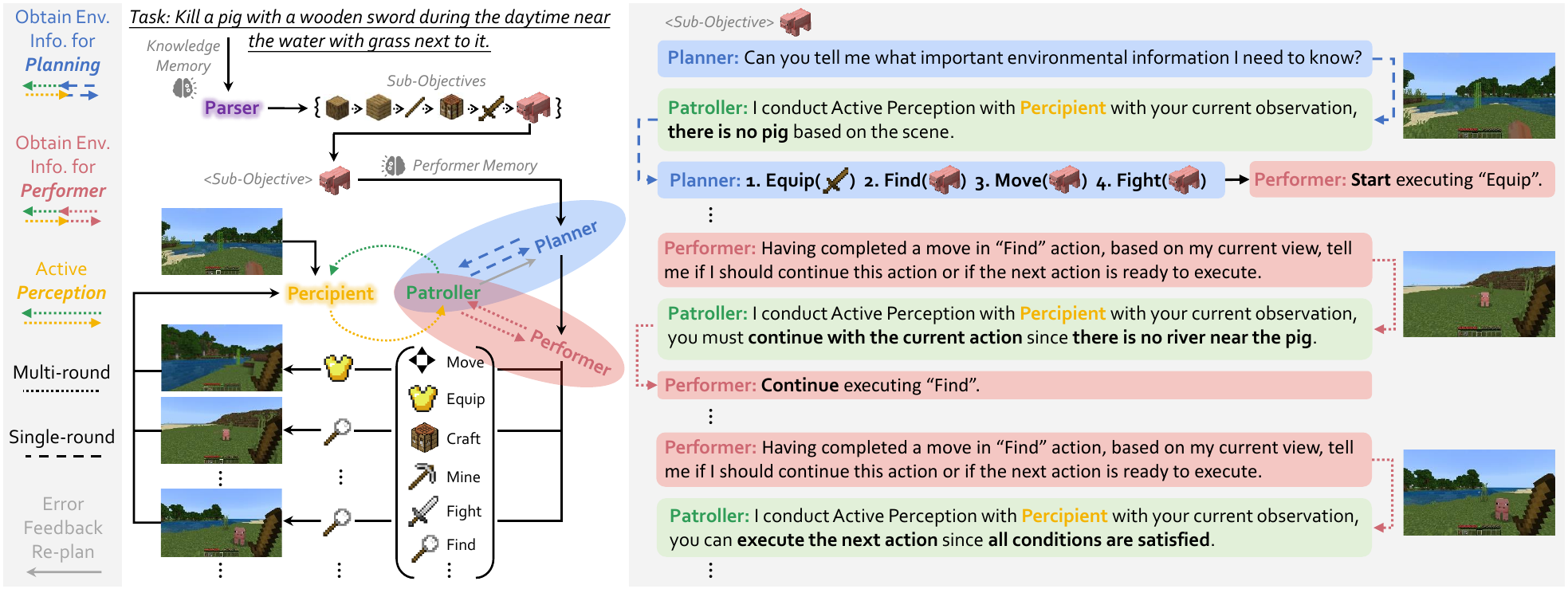}
   \caption{Overview of module interaction in {\mname}. After receiving the task instruction, {\mname} first utilizes Parser to generate a sub-objective list. Once a sub-objective is passed to the Planner, the Planner Obtaining Env. Info. for Perception-aware Planning. The performer takes frequently  Perception-aware Execution to interact with the environment by interacting with the Patroller. Both Perception-aware Planning and Execution rely on the Active Perception between the Percipient and the Patroller. Once there are execution failures, the Planner will re-schedule the action sequence of the current sub-objective. Mechanisms for collaboration and inspection of multiple modules guarantee the correctness and robustness when {\mname} is solving an open-ended embodied task.}
\label{fig:pipeline}
\vspace{-5mm}
\end{figure*}

\section{Related Work}
\label{sec:related work}

\subsection{Multi-modal Large Language Models}

With the development of Large Language Models (LLMs) like the GPT series~\cite{radford2019language,brown2020language,ouyang2022training}, as well as open-source LLMs such as the LLaMA series~\cite{touvron2023llama,touvron2023llama2} and Vicuna~\cite{chiang2023vicuna}, Multi-modal Large Language Models (MLLMs) have emerged. Examples of such MLLMs include LLaVA~\cite{liu2023visual}, InstructBLIP~\cite{instructblip}, and LAMM~\cite{yin2023lamm}, among others~\cite{zhu2023minigpt,ye2023mplug,gao2023llama,chen2023shikra,peng2023kosmos,huang2023smartedit}. In this work, we introduce MineLLM, which is specifically designed and trained for Minecraft, and leverage its perception, interaction, and analysis capabilities to build up Percipient for {\mname}, and further enable an objective-conditioned active perception scheme.

\subsection{Agents in Minecraft}


Previous works\cite{guss2019minerl,fan2022minedojo,yuan2023plan4mc,cai2023open,lin2021juewu,ding2023clip4mc,oh2017zero, zhou2024minedreamer} attempt to use approaches such as hierarchical RL, goal-based RL, and reward shaping to train an agent in Minecraft. MineCLIP~\cite{fan2022minedojo} enables the resolution of various open-ended tasks specified in free language, even without any manually designed dense rewards. DreamerV3~\cite{hafner2023mastering} succeeds in training agents in Minecraft with a learned world model. VPT~\cite{baker2022video} builds a foundation model for Minecraft by learning from massive videos. Based on VPT,  Steve-1~\cite{lifshitz2023steve} also explores bringing in MineCLIP~\cite{fan2022minedojo} to get an instruction following policy with high performance. 
%
%
The development of recent large language model-related work Voyager~\cite{wang2023voyager}, DEPS~\cite{wang2023describe}, GITM~\cite{zhu2023ghost} further promote the advancement of agents in long-horizon tasks. These works use pre-trained large language models as the zero-shot planners\cite{huang2022language} for agents, leveraging the powerful reasoning capabilities of large language models to obtain continuous operation instructions or executable policy lists.

We take advantage of the reasoning capability of LLM to build up our own agent. Existing LLM agents~\cite{zhu2023ghost, zhu2023ghost} in Minecraft feed scene data from simulation platforms~\cite{fan2022minedojo,guss2019minerl} into large language models for task planning. However, for embodied agents in real scenes, it is clearly unrealistic to use accurate scene data directly. Therefore, agents need to be robust to make decision corrections despite inaccurate or erroneous perception information. Moreover, open-ended tasks need hierarchical reasoning~\cite{oh2017zero} and complex open-ended context understanding~\cite{baker2022video,fan2022minedojo}, classical perception networks can only output fixed perception results and cannot provide corresponding perception information according to the task, making it impossible to understand open-ended scenarios. Therefore, we design \mname, an embodied agent with open-ended capabilities that can solve the problem of open-ended tasks.

\vspace{-5mm}
\section{Method}
\label{sec:Method}

In this section, we first give an overview of our proposed {\mname}, 
%
%
for solving context-dependent and process-dependent tasks in an open-world and embodied environment, such as Minecraft (\cref{sub:overview}). 
Next, we elaborate on how to implement an active perception scheme (\cref{sub:objective_conditioned_perception}).
This scheme plays a vital role in {\mname} to solve context-dependent tasks, since it reliably grounds the visual content according to different kinds of objectives, and thus strengthens the planning skill and execution correctness with respect to context-dependent tasks.
%
%
Then, we show how to plan and update action sequences in awareness of the situations, and how to reliably execute these actions in an embodied environment (\cref{sub:situation_aware_planning_and_embodied_action_execution}).
Finally, we give necessary implementation details about {\mname} in \cref{sub:implementation_details}.

\subsection{Overview}
\label{sub:overview}

As demonstrated in \cref{fig:pipeline}, our {\mname} includes five major modules, \ie, Parser, Percipient, Planner, Performer, and Patroller. 
%
%
Specifically, Percipient is a parameter-efficiently fine-tuned Multimodal Large Language Model (MLLM) that is specified to the Minecraft environment.
The Parser, Planner, and Patroller are pre-trained Large-language Models (LLMs).
We also include retrieval-augmented generation (RAG) to enhance the quality of responses generated by Parser and Planner.
Performer is an interface that explains each action from the action sequence into executable commands that directly control the game character.

\begin{figure*}[t]
\centering
\includegraphics[width=1\linewidth]{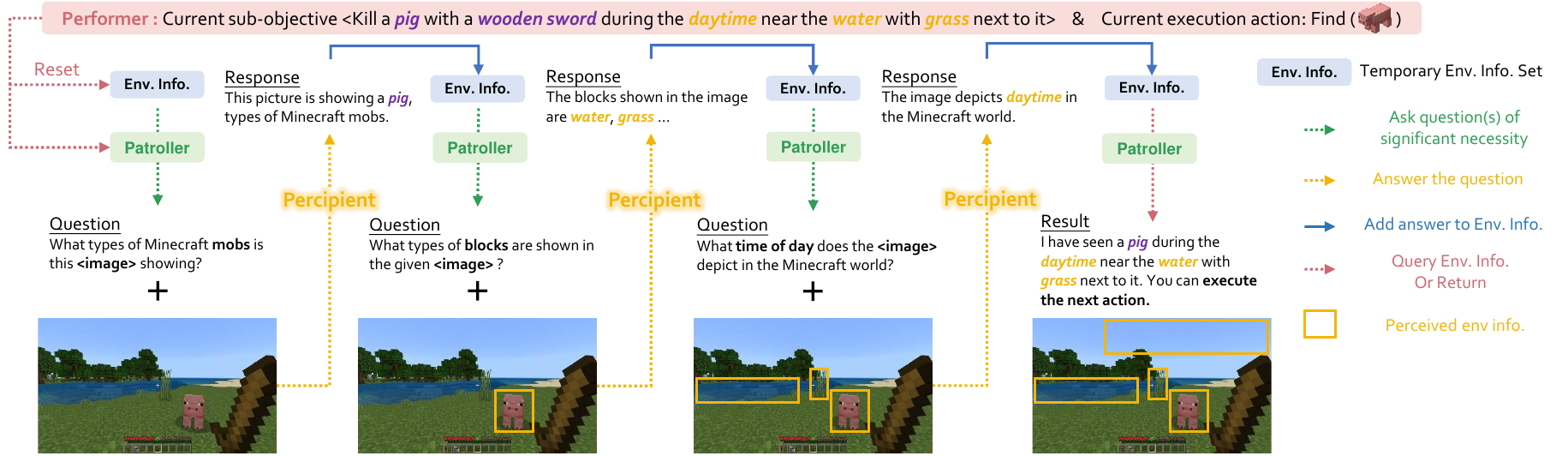}
\vspace{-5mm}
   \caption{A demonstration of the process of Active Perception scheme. Temporary Env. Info. Set saves information collected in the current scenario, so it should be reset at the beginning of Active Perception scheme. Performer then invokes Patroller to start asking Percipient questions with respect to the description of the sub-objective and the current execution action round by round. The responses of Percipient are saved in Temporary Env. Info. Set and are also gathered as the context for the next question-answering round. After finishing asking all significant necessary questions, Patroller will check whether the current execution action is complete by analyzing the current sub-objective with Perceived env info. saved in Temporary Env. Info. Set, therefore complex Context-Dependent Tasks could be solved smoothly.}
\vspace{-5mm}
\label{fig:Active perception}
\end{figure*}

\vspace{+1mm}
\noindent\textbf{Why can {\mname} solve context-dependent and process-dependent tasks?}
{\mname} includes an \emph{active perception} scheme by means of multi-round interactions between Percipient and Patroller, which is to actively perceive the environmental information in the observed images, with respect to various objectives raised by Planner or Performer.
With the help of this scheme, Planner can schedule or update action sequences in awareness of the observed images, inventory status and \etc, resulting in a \emph{situation-aware planning}; Performer can execute actions that are adapted to the embodied environment, resulting in a \emph{embodied action execution}. 
Patroller in this scheme can also feedback on better choices of plans/actions based on the visual evidence so that the process-dependent tasks are solved with fewer chances of context-dependent execution failures.
Moreover, Percipient can understand open-ended visual concepts, therefore it allows {\mname} to solve tasks that are never seen before.
%

\vspace{+1mm}
\noindent\textbf{How does {\mname} function?}
In \cref{fig:pipeline}, upon receiving a high-level task, {\mname} first utilizes the Parser to generate a sequence of short-horizon sub-objectives, as a list of rich instructions in natural languages.
The feasibility of the generated sub-objectives is augmented by retrieving an external Knowledge Memory. This knowledge mainly comes from three sources: part of it is from the online wiki, another part is from the crafting recipes of items in MineDojo~\cite{fan2022minedojo}, and some are user tips from Reddit.
To one sub-objective, Planner schedules the action sequence that is grounded by the environmental information gathered by the active perception scheme.
%
%
In this case, Performer will execute the actual actions by explaining the action sequence that is adapted to the embodied environment, via frequent interaction with the active perception scheme. 
Once there are execution failures (determined by Patroller), Planner will re-schedule the action sequence of the current sub-objective, or even update the following sub-objectives if some necessary sub-objectives are missing.
Otherwise, the agent will go to the next sub-objective and schedule new action sequences, whilst the successful action sequence of the current sub-objective will be stored in the external memory of Planner (called Performer Memory), along with the agent situation when it was planned. 
In the end, the agent will stop when the final sub-objective of the task has been reached.

\subsection{Active Perception}
\label{sub:objective_conditioned_perception}

Let's take the example shown in \cref{fig:Active perception} to demonstrate how the active perception scheme works. 
In this example, the active perception scheme is communicated with Performer to enable an embodied action execution.

At first, Performer invokes Patroller to start asking Percipient questions with respect to the description of the sub-objective and the current execution action, while simultaneously resetting the set of environmental information to be gathered.
Then Patroller progressively asks Percipient whether the observed image contains necessary items/factors (\eg, mobs~\raisebox{-0.3ex}{\includegraphics[width=0.35cm]{icon/pig.png}}~\raisebox{-0.3ex}{\includegraphics[width=0.3cm]{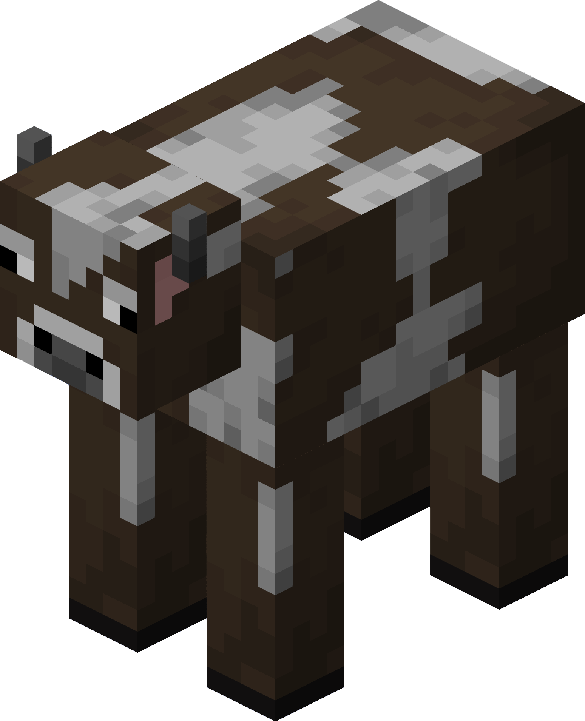}}~\raisebox{-0.3ex}{\includegraphics[width=0.3cm]{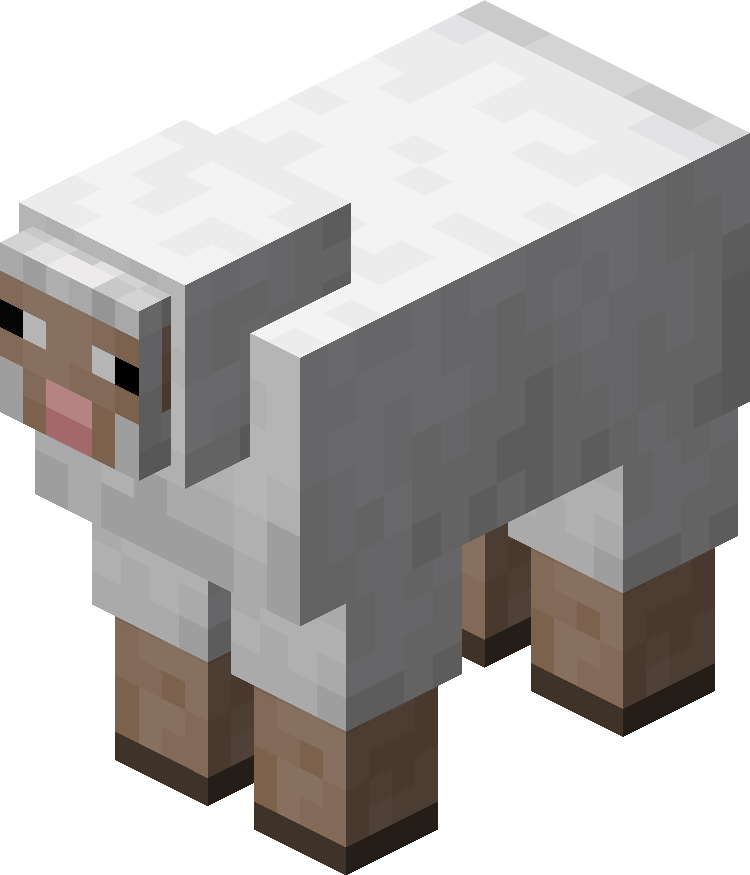}}, blocks~\raisebox{-0.3ex}{\includegraphics[width=0.3cm]{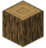}}\raisebox{-0.3ex}{\includegraphics[width=0.3cm]{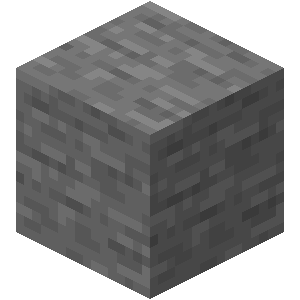}}\raisebox{-0.3ex}{\includegraphics[width=0.3cm]{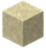}}, time~\raisebox{-0.3ex}{\includegraphics[width=0.3cm]{icon/sun.png}}\raisebox{-0.3ex}{\includegraphics[width=0.3cm]{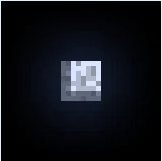}}) related to recent sub-objective (\eg, pig \raisebox{-0.3ex}{\includegraphics[width=0.35cm]{icon/pig.png}}) and the executing action (\eg, ``find pig \raisebox{-0.3ex}{\includegraphics[width=0.35cm]{icon/pig.png}}").
The responses of Percipient are also progressively gathered and act as the context for the next question-answering round.
Note that in each round, Patroller also checks whether all the necessary items/factors have been collected - If yes, Patroller stops the interaction and returns all the environmental information as natural language, and invokes Performer to execute the next action.
If Patroller eventually fails to gather enough items/factors, it will tell Performer what items/factors are missing in the observed images, which suggests Performer keeps executing the current action. Please also check the example shown in \cref{fig:pipeline}.

Similarly, active perception used in situation-aware planning is similar to what is explained here, except that the applied instructions do not contain the executable action. For more details please check the Sup.~E.

\subsection{Perception-aware Planning and Execution}
\label{sub:situation_aware_planning_and_embodied_action_execution}

\textbf{Situation-aware Planning.}
Given one sub-objective, Planner will generate the action sequence based on the description of the situation, such as the objective-conditioned environmental information from the active perception scheme, the inventory status and localization, and \etc.
Moreover, Planner will retrieve previous successful action sequences as the demonstration prompt to augment the aforementioned planning results.
If the active perception scheme fails to find the key items/factors about the current sub-objective in the observed image, the generated action sequences will include more actions to reach them.
Moreover, if Performer encounters execution failures determined by Patroller (such as failure of ``equip wooden sword \raisebox{-0.3ex}{\includegraphics[width=0.3cm]{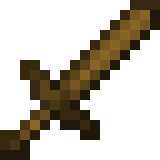}}''), Planner will re-schedule the action sequence or even update the following sub-objectives, with the help of external memories.

\vspace{+1mm}
\noindent\textbf{Embodied Action Perception.}
As indicated in \cref{sub:objective_conditioned_perception}, Performer would like to communicate with the active perception scheme in every round of action execution, so as to enhance the ego-centric awareness of the agent. 
The new action will be executed if Patroller identifies necessary environmental information in the observed images that matches both the sub-objective and the goal of the current action. Otherwise, the current action is kept executing until encountering execution failures or the end of the episode.
The successful action sequence about one sub-objective will be stored in the Performer Memory, together with necessary situational information of the agent when it was planned. For more details about the planning and execution process, please check Sup.~G.2 and Sup.~B.2.


\subsection{Implementation Details}
\label{sub:implementation_details}

\begin{figure}[t]
    \centering
    \includegraphics[width=0.7\linewidth]{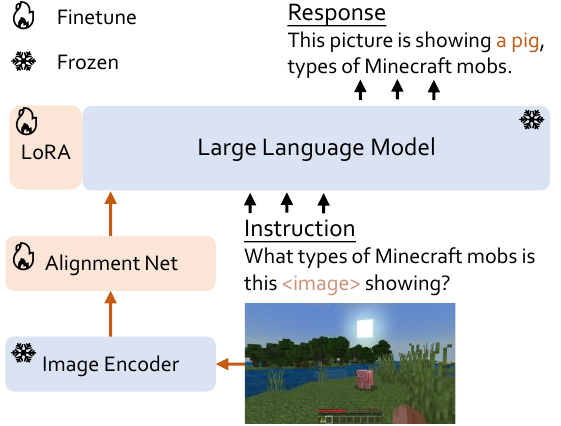}
    \caption{The model architecture of MineLLM. Image is encoded by a pre-trained vision encoder and decoded by LLM. Only the parameters of Alignment Net and LoRA are trainable.}
    \label{fig:mllm}
    \vspace{-5mm}
\end{figure} 

\noindent\textbf{Percipient.}
The network of Percipient is depicted in \cref{fig:mllm}. 
Images are processed by a frozen vision encoder MineCLIP~\cite{fan2022minedojo}, whose features are projected by an Alignment Net(we use two-layer MLP like LLaVA-1.5~\cite{liu2023improved}) to the same feature space as the text embeddings of the applied LLM (we use Vicuna-13B-v1.5~\cite{chiang2023vicuna}). 
%
%
Then the vision and text tokens are concatenated to feed into a LoRA-based fine-tuned LLM~\cite{hu2021lora}.
We add LoRA~\cite{hu2021lora} parameters to all projection layers of the self-attention layers in the LLM. Only the parameters of the Alignment Net and the LoRA module are optimized during training. 
%
%
%
The construction of the training data with respect to Percipient is in the Sup.~B.1.
%
%

\vspace{+1mm}
\noindent\textbf{Parser, Planner, and Patroller.} 
We utilize OpenAI's GPT-4~\cite{openai2023gpt} as LLMs in Parser, Patroller, and Planner.
%
%
%
We also evaluate other alternatives of GPT-4~\cite{openai2023gpt}, such as open-source models like Vicuna-13B-v1.5~\cite{chiang2023vicuna} and LLaMA2-70B-Chat~\cite{touvron2023llama2} in Sup.D.3.

\vspace{+1mm}
\noindent\textbf{Performer.}
It is important to clarify that the actions generated by Planner are not low-level commands such as keyboard and mouse operations~\cite{baker2022video}, but a set of simple actions (such as equip, move, craft).
Inspired by GITM~\cite{zhu2023ghost}, we implement these actions appropriately through basic operations provided by the MineDojo~\cite{fan2022minedojo} simulator. For more details, please check the Sup.~B.2.

\vspace{-2mm}
\section{Experiments}

%
At first, we depict the setup of the Minecraft simulation environment that we build and validate {\mname}, and give the definition of the evaluated tasks and how to set them in \cref{sec:Experimental Setup}.
In \cref{sec:Main Results}, we present the quantitative and qualitative performance of {\mname}, as well as in-depth discussions on these tasks, and demonstrate that {\mname} can even successfully accomplish tasks that are more open-ended and never seen before.
At last, we investigate how different modules affect the performance of {\mname} and analyze the impact of various module choices within our system in Sec.~\ref{sec:Ablation Study}.

%
%
%
%
%


\vspace{-1mm}
\subsection{Experimental Setup}
\label{sec:Experimental Setup}

\noindent\textbf{Environment Setting.}
We employ MineDojo~\cite{fan2022minedojo} as our simulation environment to build and validate {\mname}.
We capture player ego-view images provided by MineDojo~\cite{fan2022minedojo} as input of {\mname}, and further construct a dataset for training MineLLM. As for the output of {\mname}, we encapsulate MineDojo's~\cite{fan2022minedojo} actions to create our own action space.


\vspace{+1mm}
\noindent\textbf{Task Setting.}
To evaluate how our {\mname} can integrate perception information with planning and execution, we define two types of tasks: \textit{Context-Dependent Tasks} and \textit{Process-Dependent Tasks} as illustrated in Tab.~\ref{tab:Context-Dependent Tasks Definition in Main Paper} and Tab.~\ref{tab:Process-Dependent Tasks Definition in Main Paper}.

{\textit{1) Context-Dependent Tasks}}
primarily study how Active Perception enables the agent to better perceive low-level context information in the environment. We first establish $6$ aspects of environmental information derived from the Minecraft game environment:  \textsl{[Object, Mob, Ecology, Time, Weather, Brightness]}. Each aspect has multiple options. For example, pigs~\raisebox{-0.3ex}{\includegraphics[width=0.35cm]{icon/pig.png}}, cows~\raisebox{-0.3ex}{\includegraphics[width=0.3cm]{icon/cow.png}}, and sheep~\raisebox{-0.3ex}{\includegraphics[width=0.3cm]{icon/sheep.png}} are all elements belonging to Mob. Based on this, we define $16$ tasks and organize their difficulty into four levels by taking into account the number of information elements that require perception, as is shown in Tab.~\ref{tab:Context-Dependent Tasks Definition in Main Paper}. For example, Easy tasks necessitate the perception of only one element, whereas Complex tasks involve the perception of $4$ to $6$ elements. We rigorously assess {\mname}'s proficiency in environmental context perception across these $16$ tasks.
In \textit{Context-Dependent Tasks}, our environment details are predetermined (\eg, biomes~\raisebox{-0.3ex}{\includegraphics[width=1cm]{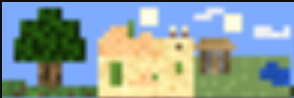}}, weather \raisebox{-0.3ex}{\includegraphics[width=1cm]{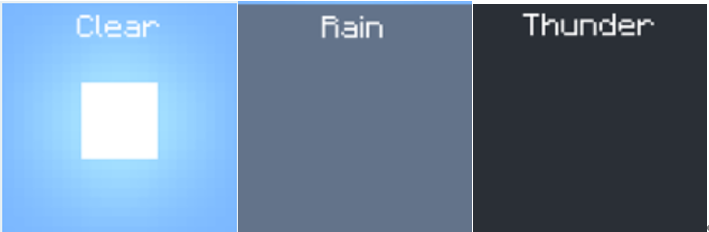}}, and \etc), as certain targets are exclusive to specific environments. Without this environmental specificity, the agent might never encounter the intended target. We retain each observation of active perception throughout the task, using them as references to ascertain the agent's successful completion of the task.

{\textit{2) Process-Dependent Tasks}}
focus on exploring the contributions of situation-aware planning, embodied action execution, and the integration with Active Perception in accomplishing long-term tasks while constantly perceiving the environment and dynamically adjusting actions.
We select $25$ tasks from the technology tree and define their difficulty levels as Basic level~\raisebox{-0.3ex}{\includegraphics[width=0.3cm]{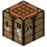}} to Diamond level~\raisebox{-0.3ex}{\includegraphics[width=0.3cm]{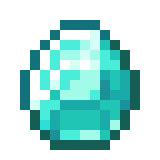}} based on the number of reasoning steps required to complete the tasks. All environmental factors (\eg, biomes~\raisebox{-0.3ex}{\includegraphics[width=1cm]{icon/biome.png}}, weather \raisebox{-0.3ex}{\includegraphics[width=1cm]{icon/weather.png}}, and \etc) are randomized in \textit{Process-Dependent Tasks}. More details can be found in Sup.D.1.

\begin{table}[t]
\caption{\textit{Context-Dependent Tasks}. $16$ tasks are defined and divided into $4$ difficulty levels based on the minimum number of information types needed. \underline{Underlines} label the environmental information, reflecting the complexity varies at each level. }
\vspace{-1mm}
\centering
\small
\begin{tabular}{cc}
\bottomrule[1pt]
\hline
Task Level & Example Task \\ 
\hline
Easy   & Find a \underline{tree} \raisebox{-0.3ex}{\includegraphics[width=0.3cm]{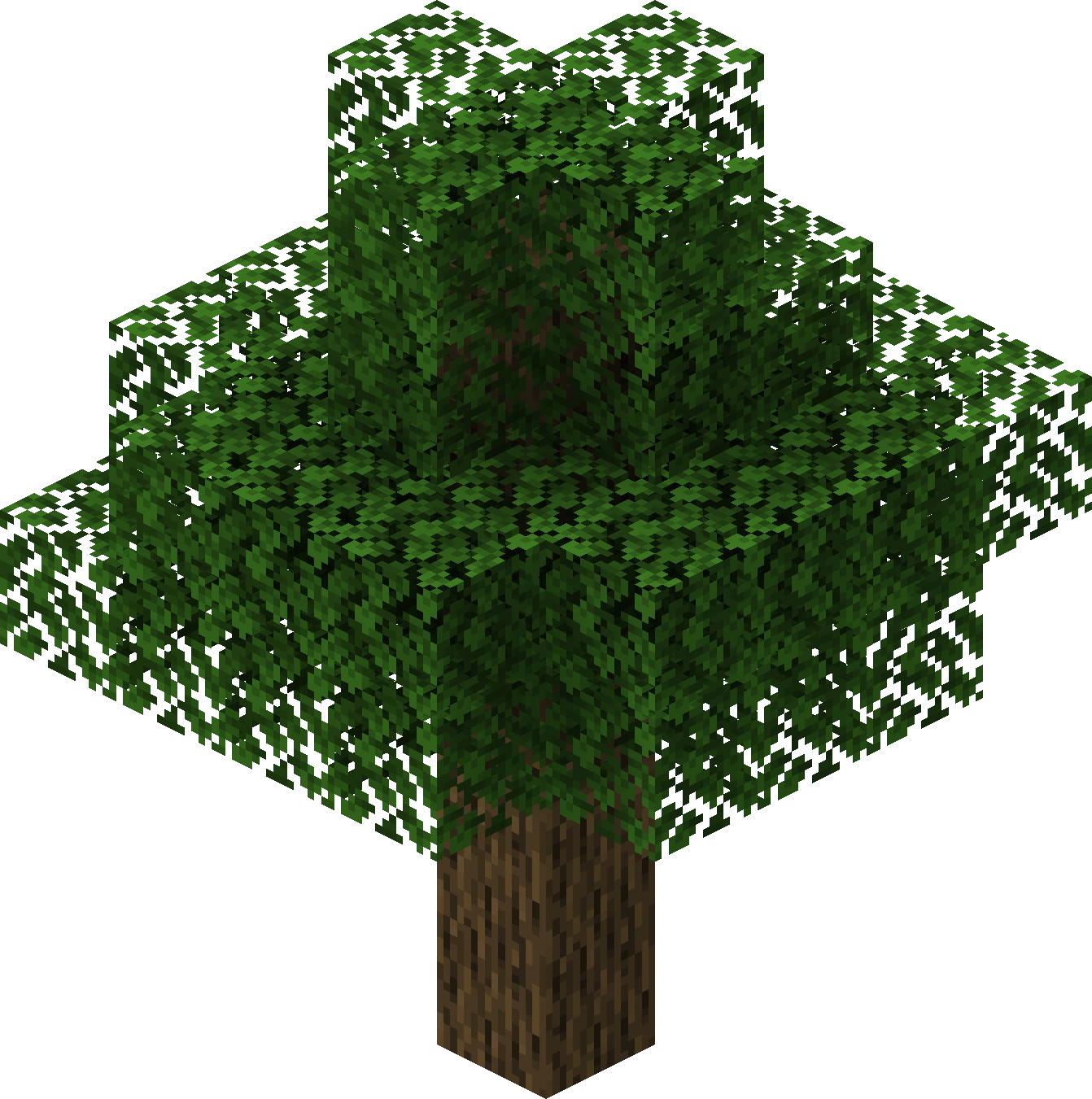}}   \\
Mid  &Find a \underline{tree} \raisebox{-0.3ex}{\includegraphics[width=0.3cm]{icon/tree.png}} in the \underline{forest} ~\raisebox{-0.3ex}{\includegraphics[width=0.3cm]{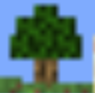}}    \\
\multirow{2}{*}{Hard}    & Find a \underline{tree} \raisebox{-0.3ex}{\includegraphics[width=0.3cm]{icon/tree.png}} in the \underline{forest} ~\raisebox{-0.3ex}{\includegraphics[width=0.3cm]{icon/forest.png}}  \\ 
 &  during the \underline{nighttime}  \raisebox{-0.3ex}{\includegraphics[width=0.3cm]{icon/moon.png}}\\
\multirow{2}{*}{Complex}   & Find a \underline{pig} \raisebox{-0.3ex}{\includegraphics[width=0.35cm]{icon/pig.png}} near a \underline{grass} \raisebox{-0.3ex}{\includegraphics[width=0.3cm]{icon/grass.png}} \\
                        & in the \underline{forest} ~\raisebox{-0.3ex}{\includegraphics[width=0.3cm]{icon/forest.png}} during the \underline{daytime}  \raisebox{-0.3ex}{\includegraphics[width=0.3cm]{icon/sun.png}} \\ 
\bottomrule[1pt]
\end{tabular}
\vspace{-5mm}
\label{tab:Context-Dependent Tasks Definition in Main Paper}
\end{table}

\noindent\textbf{Evaluation Metrics.}
For different tasks, the agent's initial position and environment seed are randomized. The agent begins in survival mode, commencing with an empty inventory, and faces the challenge of hostile mob generation. It starts from scratch, with a game time limit of 10 minutes, a time period equivalent to 12,000 steps at a control frequency of 20Hz. More details can be found in Sup.~C.

For the \textit{Context-Dependency Tasks}, each assignment is open-ended. Therefore, we conduct manual evaluations when the agent determines it has completed the task or exceeds the time limit. Two cases are ruled as failures: 1)There is an observation that meets all the conditions, but the agent does not end the task; 2) The last observation does not meet all the conditions, yet the agent ends the task. Otherwise, we believe that the agent correctly perceives all the context according to the task and determines that the task is successfully completed. For the \textit{Process-Dependent Tasks}, any accidental deaths of the agent during the game are counted as failures, as are instances where the agent does not accomplish the task within the time limit. 

In practice, we conduct $50$ games on \textit{Context-Dependent Tasks} and $30$ games on \textit{Process-Dependent Tasks}, averaging the success rates for both. The results are grouped according to the previously defined difficulty levels, and report the group means.
For detailed definitions of the evaluation, please refer to Sup.~D.

\begin{table}[t]
\caption{\textit{Process-Dependent Tasks}. $25$ tasks are defined and divided into $5$ difficulty levels based on incrementally increasing reasoning steps. A higher difficulty level implies that the agent needs to engage in longer reasoning and planning with the environment.}
\vspace{-1mm}
\centering
\small
\begin{tabular}{ccc}
\bottomrule[1pt]
Task Level& Reasoning Step & Example Task \\ 
\hline
Basic \raisebox{-0.3ex}{\includegraphics[width=0.3cm]{icon/crafting_table.png}}    & 1-3   & craft crafting table  \raisebox{-0.3ex}{\includegraphics[width=0.3cm]{icon/crafting_table.png}}  \\
Wooden \raisebox{-0.3ex}{\includegraphics[width=0.3cm]{icon/wooden_sword.png}}    & 4-5   & craft wooden sword  \raisebox{-0.3ex}{\includegraphics[width=0.3cm]{icon/wooden_sword.png}}   \\ 
Stone \raisebox{-0.3ex}{\includegraphics[width=0.3cm]{icon/stone.png}}    & 6-9  & mine stone \raisebox{-0.3ex}{\includegraphics[width=0.3cm]{icon/stone.png}} \\ 
Iron \raisebox{-0.3ex}{\includegraphics[width=0.3cm]{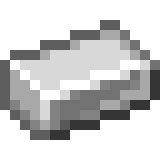}}   & 10-11  & smelt iron ingot  \raisebox{-0.3ex}{\includegraphics[width=0.3cm]{icon/iron.png}} \\ 
Diamond \raisebox{-0.3ex}{\includegraphics[width=0.3cm]{icon/diamond.png}}    & \textgreater11  & obtain diamond  \raisebox{-0.3ex}{\includegraphics[width=0.3cm]{icon/diamond.png}} \\ 
\bottomrule[1pt]
\end{tabular}
\vspace{-5mm}
\label{tab:Process-Dependent Tasks Definition in Main Paper}
\end{table}

\subsection{Main Results}
\label{sec:Main Results}

\subsubsection{Results of Context-Dependent Tasks}
\label{sec:Context-Dependent Tasks Results}

In \textit{Context-Dependent Tasks}, we primarily investigate how to enhance an agent's perception of context information within the environment. We demonstrate the performance difference between Active Perception and other perception methods.
We compare them with pre-trained multi-modal large language models LLaVA-1.5~\cite{liu2023improved} and GPT-4V~\cite{gpt4v}, and analyze the performance of both active and fine-grained global perception on the tasks in \cref{tab:main result context}. Although fine-grained global perception can obtain comprehensive perceptual information, due to the lack of objective-conditioned attention, the objective-related information obtained may be lacking or incorrect. Active perception only focuses on objective-related information and ignores other useless information, so that more accurate objective-related information can be obtained and better performance in \textit{Context-Dependent Tasks} can be achieved.
For the comparison, we use MineLLM, which is fine-tuned on the Minecraft instruction dataset we collect, slightly better than GPT-4V~\cite{gpt4v}, which is trained on massive data, and substantially better than LLaVA-1.5~\cite{liu2023improved}, which is not fine-tuned on instruction data. The complete results of \textit{Context-Dependent Tasks} can be found in Sup.D.2.

    
    


\vspace{-3mm}
\subsubsection{Results of Process-Dependent Tasks}
\label{sec:Process-Dependent Tasks Results}

In \textit{Process-Dependent Tasks}, we report the performance of the agent in completing long-horizon tasks by continuously perceiving the environment context and dynamically adjusting its actions. We also investigate the agent's behavior in scenarios of non-situation-aware planning and non-embodied action execution. The complete results of \textit{Process-Dependent Tasks} can be found in Sup.D.2.

 In considering the landscape of related works~\cite{zhu2023ghost,wang2023describe,wang2023voyager,baker2022video,hafner2023mastering}, we refrain from making direct comparisons due to the substantial variations in the \textbf{observation space}, \textbf{action space}, \textbf{environmental setup}, and \textbf{game termination conditions}. Notably, VPT~\cite{baker2022video} emulates human players' keyboard and mouse controls, DreamerV3~\cite{hafner2023mastering} is trained from scratch for diamond collection \raisebox{-0.3ex}{\includegraphics[width=0.3cm]{icon/diamond.png}} in a modified Minecraft environment with altered block-breaking mechanics using world models, DEPS~\cite{wang2023describe} integrates LLM planning and a learning-based control policy based on MineDojo~\cite{fan2022minedojo} actions, GITM~\cite{zhu2023ghost} employs privileged information such as lidar perception, and Voyager~\cite{wang2023voyager} utilizes purely text-based information perception in collaboration with the Mineflayer API for action. Given that our experiments aim to showcase the system's capability to adapt both process-dependent reasoning and complex context-understanding tasks, our focus turns to presenting two key insights drawn from the system's performance, as detailed below.

\begin{table}[t]
\caption{Performance on \textit{Context-Dependent Tasks}. We compare the success rate of different Methods and different Perception strategies. We set up special prompt to make the output of the caption as comprehensive as possible, this perception method is called Fine-Grained Global Perception. We use A to denote Active Perception, and G to denote Fine-Grained Global Perception.}
\vspace{-1mm}
\centering
\footnotesize
\begin{tabular}{lc|ccccccc}
\bottomrule[1pt]
\multirow{2}{*}{Method} &\multirow{2}{*}{Strategy} &\multicolumn{4}{c}{Average Success Rate(\%)}  \\
                              &  & Easy & Mid & Hard & Complex \\
\hline
 \multirow{2}{*}{LLaVA-1.5~\cite{liu2023improved}}     & G  &47.5 & 22.5 & 5.0 & 0.0    \\
    & A  &72.5 & 50.0 & 11.0 & 0.0    \\
\hline
 \multirow{2}{*}{GPT-4V~\cite{gpt4v}}     & G  & 97.5 & 85.0 & 75.0 & 60.0    \\
  & A & \textbf{100.0} & 94.5 & 92.5 & 87.5  \\
\hline
 \multirow{2}{*}{\textbf{\mname(Ours)}}    & G & 90.0 & 82.5 & 77.5 & 67.5 \\
   & A & \cellcolor{gray!15}98.5 & \cellcolor{gray!15}\textbf{94.5} & \cellcolor{gray!15}\textbf{93.0} & \cellcolor{gray!15}\textbf{91.0} \\

\arrayrulecolor{black}
\bottomrule[1pt]
\end{tabular}
\label{tab:main result context}
\end{table}

\vspace{+1mm}
\noindent\textbf{Embodied action execution is critical for open-ended tasks.} 
Comparing {\mname} w/o E. and {\mname} in Tab.~\ref{tab:Process-Dependent Tasks Result in Main Paper}, we can observe that when an agent is unable to interact with the environment and access low-level environment contextual information during action execution, it essentially becomes ``blind'',  unable to determine the termination of its actions based on environment. Therefore, the success rate in \textit{Process-Dependent Tasks} is $0.00\%$.

\vspace{+1mm}  
\noindent\textbf{Situation-aware planning leads to more scenario-appropriate strategies.} 
Comparing {\mname} w/o P. and {\mname} in Tab.~\ref{tab:Process-Dependent Tasks Result in Main Paper}, we observe that the lack of environment contextual information during the agent's planning process can lead to erroneous or redundant actions, thereby reducing the success rate (for example, the success rate in diamond-level \raisebox{-0.3ex}{\includegraphics[width=0.3cm]{icon/diamond.png}} tasks decrease from $22.00\%$ to $14.00\%$). Consider a scenario where the current sub-objective is ``kill a pig \raisebox{-0.3ex}{\includegraphics[width=0.3cm]{icon/diamond.png}}''. If a pig \raisebox{-0.3ex}{\includegraphics[width=0.35cm]{icon/pig.png}} is already present, the agent should directly execute ``move'' to approach without the need to first ``find'' then ``move''. However, the relatively small decrease in the success rate can be attributed to the dynamic adjustment of perception and action execution offered by embodied action execution. Simultaneously, when errors are detected, the perceived environmental information and the erroneous actions can be fed back to the planner for re-planning.

\subsubsection{Open-Ended Tasks}
\label{sub:open_ended_tasks}

\begin{figure}[t]
    \centering
    \includegraphics[width=1\linewidth]{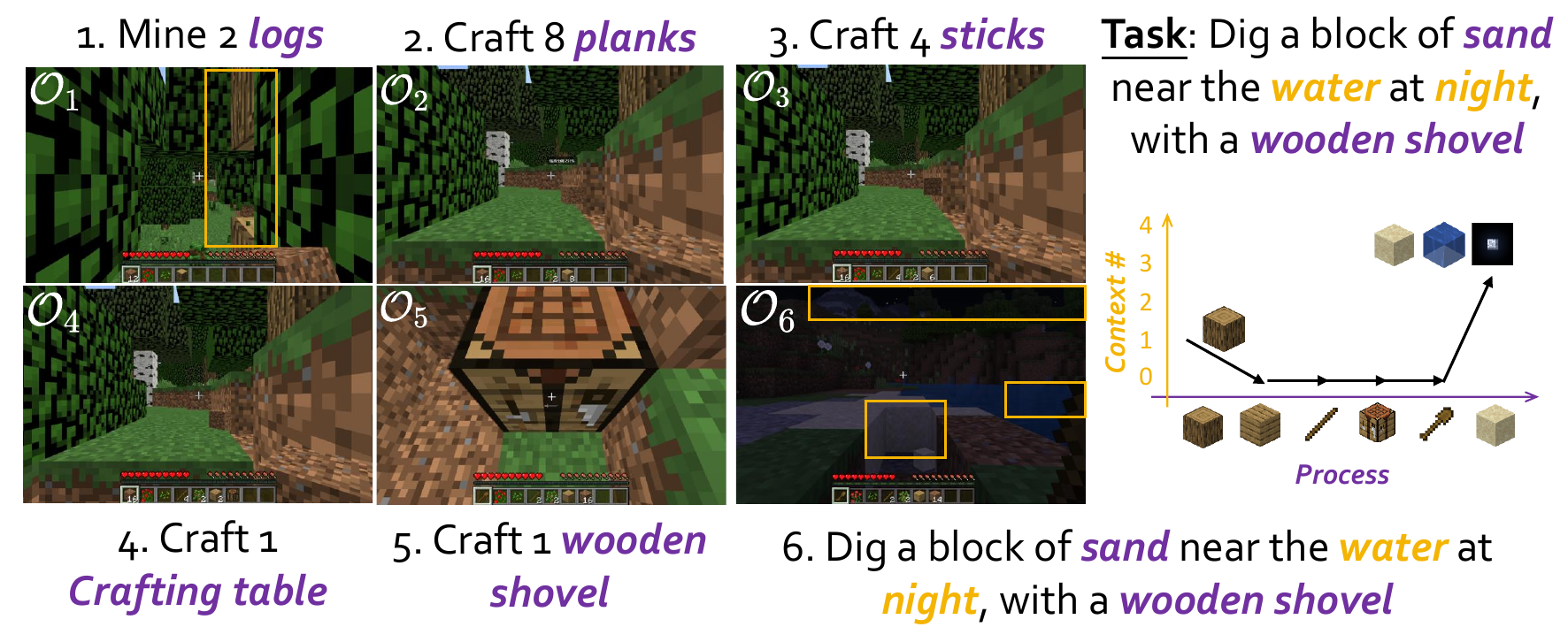}
    \caption{Screenshots of \textit{``Dig a block of sand \raisebox{-0.3ex}{\includegraphics[width=0.3cm]{icon/sand.png}} near the water \raisebox{-0.3ex}{\includegraphics[width=0.3cm]{icon/water.png}} at night  \raisebox{-0.3ex}{\includegraphics[width=0.3cm]{icon/moon.png}} with a wooden shovel \raisebox{-0.3ex}{\includegraphics[width=0.3cm]{icon/wooden_shovel.png}}''}. In \textit{Open-Ended Tasks}, the agent needs to better integrate low-level context information and high-level decision-making, making it extremely challenging. 
    }
    \vspace{-5mm}
    \label{fig:Demo}
\end{figure}

Processing long-horizon reasoning and understanding complex contexts are interconnected in the real world. For simplicity and comparability of the experimental setup, the first two task settings do not consider the intersection of process and context, as we cannot exhaust all combinations that these two task dimensions can form.
Therefore, we refer to tasks that incorporate both Process-Dependent and Context-Dependent elements as \textit{Open-Ended Tasks}. Specifically, these tasks require the agent to perceive different information of the environment at multiple stages of completing sub-objectives.
As shown in \cref{fig:Demo}, we present an example of an \textit{Open-Ended Task}, named \textit{``Dig a block of sand \raisebox{-0.3ex}{\includegraphics[width=0.3cm]{icon/sand.png}} near the water \raisebox{-0.3ex}{\includegraphics[width=0.3cm]{icon/water.png}} at night \raisebox{-0.3ex}{\includegraphics[width=0.3cm]{icon/moon.png}} with a wooden shovel \raisebox{-0.3ex}{\includegraphics[width=0.3cm]{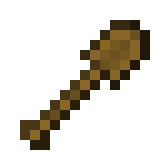}}''}.
We conduct extensive validations on this type of task, proving that {\mname} can complete long-sequential tasks in challenging environments. More demonstrations and experimental results of \textit{Open-Ended Tasks} can be found in Sup.F.3.

\begin{table}[t]
\caption{Performance on \textit{Process-Dependent Tasks}. We compare the success rate when interacting or not interacting with the environment during the planning or execution. w/o P. and w/o E. indicates non-situation-aware planning and non-embodied action execution. }
\vspace{-2mm}
\centering
\footnotesize
\begin{tabular}{l|cccccc}
\bottomrule[1pt]
\multirow{2}{*}{Method} &  \multicolumn{5}{c}{Average Success Rate(\%)} \\
                        &      Basic & Wooden & Stone & Iron & Diamond \\
\hline
 \mname~ w/o P.    &  0.00 & 0.00 & 0.00 & 0.00 & 0.00 \\  
 \mname~ w/o E.    &  92.00 & 86.00 & 68.67 & 45.33 & 14.00 \\  
\textbf{\mname}  &  96.00 & 88.67 & 76.00 & 52.00 & 22.00 \\  
\arrayrulecolor{black}
\bottomrule[1pt]
\end{tabular}
\label{tab:Process-Dependent Tasks Result in Main Paper}
\end{table}

\vspace{-2mm}
\subsection{Ablation Study}
\label{sec:Ablation Study}

We conduct ablation studies to evaluate the effectiveness of various modules. The experimental setup and the associated success rates are in \cref{sec:Experimental Setup}. More detailed ablation studies are listed in Sup.D.3. The following paragraphs present the analyses derived from our ablation studies.

\begin{table}[t]
\caption{Success rates for different MLLMs and pre-trained visual encoders in the percipient on \textit{Context-Dependent Tasks}}
\vspace{-2mm}
\centering
\scriptsize
\begin{tabular}{lc|cccc}
\hline
\multirow{2}{*}{Method} & Visual              & \multicolumn{4}{c}{Average Success Rate(\%)} \\
\multicolumn{1}{c}{}    & Encoder             & Easy & Mid & Hard & Complex\\ 
\hline
LLaVA-1.5~\cite{liu2023improved}   & CLIP~\cite{radford2021learning}            &72.50 & 50.00 & 11.00 & 0.00 \\ 
MineLLM & CLIP~\cite{radford2021learning}            &95.00 & 90.00 & 87.00 & 80.00 \\ 
MineLLM & MineCLIP~\cite{fan2022minedojo}        &98.50 & 94.50 & 93.00 & 91.00 \\ 

\hline
\end{tabular}
\label{tab:Ablation Study for mineclip in Main Paper}
\end{table}

\begin{table}[t]
\caption{Success rates for different LLMs as zero-shot Planner on \textit{Process-Dependent Tasks}}
\vspace{-2mm}
\centering
\scriptsize
\begin{tabular}{l|ccccccc}
\bottomrule[1pt]
\multirow{2}{*}{Planner} & \multicolumn{5}{c}{Average Success Rate(\%)}  \\
                                & Basic & Wooden & Stone & Iron & Diamond \\
\hline
 Vicuna-13B-v1.5~\cite{chiang2023vicuna}        & 1.33 & 0.00 & 0.00 & 0.00 & 0.00    \\
 GPT-3.5-turbo~\cite{chatgpt} & 95.33 & 86.67 & 42.00 & 2.67 & 0.00 \\
 GPT-4~\cite{openai2023gpt}         & 96.00 & 88.67 & 76.00 & 52.00 &22.00\\

\arrayrulecolor{black}
\bottomrule[1pt]
\end{tabular}
\vspace{-5mm}
\label{tab:Ablation Study Result in Main Paper gpt type}
\end{table}


\vspace{+1mm}
\noindent\textbf{Model pre-trained on massive data of Minecraft can better comprehend the Minecraft appearance styles.} 
We conduct ablation studies on the multi-modal large language model (MLLM) part within \textit{Context-Dependent Tasks} in Tab.~\ref{tab:Ablation Study for mineclip in Main Paper}, comparing the performance outcomes of different MLLMs and different pre-trained visual encoders in the percipient.
We find the performance of the open-source model LLaVA-1.5~\cite{liu2023improved} to be relatively weak, with a success rate of merely $50.00\%$ at the Mid level and $11.00\%$ on the Hard level. This is primarily due to the model's training predominantly on real-world data, causing it to struggle with the pixel-style image recognition characteristic of Minecraft. We also discover that, when the visual encoder is frozen, the MineLLM with CLIP~\cite{radford2021learning} as its visual encoder consistently performs worse across all levels compared to MineLLM with MineCLIP's~\cite{fan2022minedojo} pre-trained single image visual encoder. It may caused by, in the case of a frozen visual encoder, a visual encoder pretrained on massive data of Minecraft can align with pixel-style images more rapidly.

\vspace{+1mm}  
\noindent\textbf{Enhanced reasoning ability results in improved planning.}
We compare the performance of open-source large language models, OpenAI's GPT-3.5-turbo~\cite{chatgpt} in Tab.~\ref{tab:Ablation Study Result in Main Paper gpt type}, and GPT-4~\cite{openai2023gpt} as zero-shot Planners on \textit{Process-Dependent Tasks}. We find that as the models' inferential capabilities increase, the Planner produces better results by planning in a situation-aware method, yielding more concise and accurate execution actions. The Vicuna-13B-v1.5~\cite{chiang2023vicuna} model, when used as a Planner, struggles to produce effective plans, achieving only a $1.33\%$ accuracy rate at the Basic level~\raisebox{-0.3ex}{\includegraphics[width=0.3cm]{icon/crafting_table.png}}. GPT-4~\cite{openai2023gpt} exhibits the best performance, attaining a $22.00\%$ success rate at the Diamond level~\raisebox{-0.3ex}{\includegraphics[width=0.3cm]{icon/diamond.png}}, whereas both Vicuna-13B-v1.5~\cite{chiang2023vicuna} and GPT-3.5-turbo~\cite{chatgpt} score $0.00\%$.

\vspace{+1mm}  
\noindent\textbf{Leveraging memory leads to better planning.}
In our Performer Memory, we store previously successful sub-objectives and their corresponding execution actions. When planning in similar scenarios, Performer Memory can provide the Planner with similar execution action plans for completing the sub-objectives. While the plans may not be identical, they can effectively assist the Planner in performing situation-aware planning. Comparing the first and last rows of Tab.~\ref{tab:Ablation Study Result in Main Paper 2}, we find that without the Performer Memory, the success rate of tasks at all levels decreases (Diamond level \raisebox{-0.3ex}{\includegraphics[width=0.3cm]{icon/diamond.png}} drops from $22.00\%$ to $16.67\%$). However, the decrease is not significant as the Performer Memory primarily serves a reference function, with specific action planning still heavily reliant on the Planner's capabilities.

\vspace{+1mm}  
\noindent\textbf{Robustness is essential in open-world settings.}  
To enhance the robustness evaluation of our system, we introduce a \textit{``Random Drop''} setting. In this setting, we randomly discard one complete sub-objective from the inventory at the start of each new sub-objective, which deliberately induces execution errors for the agent. Comparing the second and third lines in Tab.~\ref{tab:Ablation Study Result in Main Paper 2},  we observe the critical role of the Patroller in recognizing feedback errors. The Patroller's ability to integrate current environmental information with error information is essential for enabling the planner to re-plan. The significance of this robustness is evident when examining the success rates. Without the Patroller's robustness, the agent's success rate on the Wooden level~\raisebox{-0.3ex}{\includegraphics[width=0.3cm]{icon/wooden_sword.png}} plummets from $76.67\%$ to $7.33\%$, while success rates on the Iron \raisebox{-0.3ex}{\includegraphics[width=0.3cm]{icon/iron.png}}, and Diamond \raisebox{-0.3ex}{\includegraphics[width=0.3cm]{icon/diamond.png}} levels drop to $0.00\%$. Details regarding the \textit{``Random Drop''} setting can be found in Sup.D.3.



\begin{table}[t]
\caption{Success rates on different modules within \textit{Process-Dependent Tasks}: We study the roles of the Performer Memory~(PM) and the check part of Patroller~(P), with 'RD' denoting \textit{``Random Drop''} setting. \checkmark denotes the inclusion of the module or setting, and \ding{55} indicates its absence. }
\vspace{-2mm}
\centering
\footnotesize
\begin{tabular}{cc|c|ccccccc}
\bottomrule[1pt]
\multirow{2}{*}{PM}  & \multirow{2}{*}{P} &  \multirow{2}{*}{RD}   & \multicolumn{5}{c}{Average Success Rate(\%)}  \\
     &                            &      & Basic & Wooden & Stone & Iron & Diamond \\
\hline
  \ding{55}        & \checkmark      & \ding{55}         & 96.00 & 87.33 & 67.33 & 47.33 & 16.67    \\
  \checkmark      & \ding{55}        &  \checkmark         & 70.00 & 7.33 & 0.67 & 0.00 & 0.00 \\
  \checkmark    & \checkmark       &  \checkmark         &  87.33 & 76.67 & 45.33 & 18.67 & 1.33\\
  \checkmark      & \checkmark       &  \ding{55}         & 96.00 & 88.67 & 76.00 & 52.00 & 22.00\\

\arrayrulecolor{black}
\bottomrule[1pt]
\end{tabular}
\vspace{-6mm}
\label{tab:Ablation Study Result in Main Paper 2}
\end{table}
\vspace{-1mm}
\section{Conclusion}
\label{sec:Conclusion}

In this paper, we propose a novel multi-modal embodied system termed \mname~which is driven by frequently ego-centric scene perception for task planning and execution.
In practice, it is designed by integrating five functional modules to accomplish task planning and execution via actively acquiring essential visual information from the scene.
The experimental results suggest that our system represents an effective integration of perception, planning, and execution, skillfully crafted to handle both context- and process-dependent tasks within an open-ended environment.


\vspace{+1mm}
\noindent\textbf{Limitation and Future Work.} 
Despite the impressive results of our approach, two major limitations need to be clarified. Firstly, the reliance on GPT-3.5-turbo~\cite{chatgpt} or GPT-4~\cite{openai2023gpt} limits the system's usability, as not everyone has access to these APIs. 
Secondly, the scope of the applied simulation platform is limited. Despite showing promising performance in Minecraft, we haven't extended our exploration to other simulation platforms, which is a potential area for further research.

\vspace{+1mm}
\noindent\textbf{Acknowledgement.}
This work was supported by the National Key R\&D Program of China (2021YFB1714300), the National Natural Science Foundation of China (62106154, 62132001), the Natural Science Foundation of Guangdong Province, China (2022A1515011524).
{
    \small
    \bibliographystyle{ieeenat_fullname}
    \bibliography{main}
}

\appendix
\clearpage
\setcounter{page}{1}
\maketitlesupplementary


\noindent The supplementary document is organized as follows:
\newline
\begin{flushleft}

\begin{itemize}
    \item Sec.~\ref{sup:Discussion}: Discussion of {\mname}.
    \newline
    \item Sec.~\ref{sup:Implementation Details}: Percipient, Memory, Observation Space and Action Space.
    \newline
    \item Sec.~\ref{sup:Environment Setting}: Environment Setting.
    \newline
    \item Sec.~\ref{sup:Task Details and Experiment Results}: Task Details, Sucess Rates of All Tasks and Ablation Study.
    \newline
    \item Sec.~\ref{sup:Different Strategy of Active Perception}: Different Strategy of Active Perception.
    \newline
    \item Sec.~\ref{sup:Applications}: Applications Demonstration of {\mname}.
    \newline
    \item Sec.~\ref{sup:Interactions}:  Interactions in {\mname}.
    \newline
\end{itemize}
\end{flushleft}

\begin{table*}[htbp]
\vspace{-4mm}
\caption{Explicit comparisons of setups and consequences.} 
\vspace{-4mm}
\centering

\tiny

\begin{tabular}{l||p{0.7cm}<{\centering}|c||c|c|c||c|c|c||c|c|c}
\bottomrule[1pt]
 \multirow{3}{*}{Method} & \multicolumn{2}{c||}{Observation Space}    & \multicolumn{3}{c||}{Action Space}            & Instruct & Situation-   & Situation-  & \multicolumn{3}{c}{Tasks Agents Can Perform} \\
                         & Info  & Not                &  Action     &  Action  & Primitive    & Following & aware    & aware           & Process   & Context   & Process \&  Context          \\
                         & Type          & Omniscient &  Type       &  Num     & Library Size             &      &Plan    & Execution          & (Long-Horizon Tasks)   & (High Env. Info Tasks)   &  (Combination of 2 Tasks)        \\

 \hline                         
 DreamerV3 & RGB        & \textcolor{green!70!black}{\checkmark}    & Original MineRL      & 25         & 0 & \textcolor{red}{\ding{55}} &\textcolor{red}{\ding{55}}   &\textcolor{red}{\ding{55}} &\textcolor{green!70!black}{\checkmark}      &\textcolor{red}{\ding{55}}&\textcolor{red}{\ding{55}}\\


 DEPS  & RGB        & \textcolor{green!70!black}{\checkmark}    & Compound (MineDojo)  & 42         & 0 & \textcolor{green!70!black}{\checkmark}  &\textcolor{red}{\ding{55}}  &\textcolor{red}{\ding{55}}        &\textcolor{green!70!black}{\checkmark}     &\textcolor{red}{\ding{55}}&\textcolor{red}{\ding{55}}\\

 GITM & Text & \textcolor{red}{\ding{55}}               & Manual (MineDojo)    & 9          & 9 + corner cases & \textcolor{green!70!black}{\checkmark} &\textcolor{red}{\ding{55}}  &\textcolor{red}{\ding{55}}             &\textcolor{green!70!black}{\checkmark}     &\textcolor{red}{\ding{55}} &\textcolor{red}{\ding{55}}\\

 Voyager& Text        & \textcolor{red}{\ding{55}}               & JavaScript APIs        & N/A       & All & \textcolor{green!70!black}{\checkmark}&\textcolor{red}{\ding{55}}  &\textcolor{red}{\ding{55}}                &\textcolor{green!70!black}{\checkmark}     &\textcolor{red}{\ding{55}} &\textcolor{red}{\ding{55}}\\

 \textbf{{\mname}(ours)}   & RGB         & \textcolor{green!70!black}{\checkmark}   & Compound (MineDojo)  & 10  & 4 & \textcolor{green!70!black}{\checkmark}&\textcolor{green!70!black}{\checkmark} &\textcolor{green!70!black}{\checkmark} &\textcolor{green!70!black}{\checkmark}     &\textcolor{green!70!black}{\checkmark} &\textcolor{green!70!black}{\checkmark}\\

\arrayrulecolor{black}
\bottomrule[1pt]
\end{tabular}
\label{tab:compare}
\end{table*}

\section{Discussion of \mname}
\label{sup:Discussion}
\begin{figure}[t]
    \centering
    \includegraphics[width=0.7\linewidth]{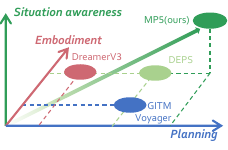}
    \caption{Paradigm innovation.}
    \label{fig:rebuttal}
\end{figure}

As shown in {\cref{fig:rebuttal}}, existing methods usually follow the paradigm that an agent should be with the ability of planning (\eg, GITM \& Voyager), embodiment (\eg, DreamerV3), or both of them (\eg, DEPS).
To be different, {\mname} introduces a new paradigm that these components (\ie, embodiment, planning, and execution as well), should be enhanced with the awareness of the situation (\ie, rich contextual and procedural information \wrt the task), which is more human-like and has the potential to solve more difficult open-end context- and process-dependent tasks (as evaluated in Sec. 4.2.3).
\textbf{(2)} Technically, {\mname} comprises five interactive modules to meet the requirement of the new paradigm, with an MLLM-based \emph{active perception} scheme to fulfil the situation awareness. To our best knowledge, {\mname} is the first embodied system in Minecraft that is capable of situation-aware planning and action execution.
\textbf{(3)} Only our method can solve Context-Dependent Tasks (in \cref{tab:compare}) by leveraging the situation-aware planning and execution. We constructed a detailed benchmark (in Sec. 4) to explore how agents complete tasks required complex reasoning and constrained by extensive environmental information.

We have provided a comparison of setups and their consequences in \cref{tab:compare}. It tells that the proposed {\mname} applies ego-centric RGB images, just utilizes a restrained set of human-defined primitives, but can solve the most challenging context-dependent and process-dependent tasks.

\section{Implementation Details}
\label{sup:Implementation Details}

\begin{table*}[]
\caption{Comparison of Observation Spaces Among Different Methods}
\centering
\begin{tabular}{lll}
\bottomrule[1pt]

Method                  & Perceptual Observation                & Status Observation        \\
\hline
\multirow{2}{*}{GITM~\cite{zhu2023ghost}}                       & LiDAR rays                            & life statistics                  \\                          
                                            & $10 \times 10 \times 10$ Voxels       & GPS, inventory, equipment        \\
\hline
\multirow{2}{*}{DreamV3~\cite{hafner2023mastering}}                    & Ego-View RGB                          & life statistics        \\
                                            &                                       & inventory, equipment        \\
\hline
\multirow{2}{*}{VPT~\cite{baker2022video}}                        & \multirow{2}{*}{Ego-View RGB}         & \multirow{2}{*}{$\varnothing$}       \\
                                            &                                       &         \\
\hline
\multirow{2}{*}{DEPS~\cite{wang2023describe}}                       & Ego-View RGB                          & Compass      \\
                                            & $3 \times 3 \times 3$ Voxels          & GPS, equipment        \\
\hline
\multirow{3}{*}{JARVIS-1~\cite{wang2023jarvis}}                   & \multirow{3}{*}{Ego-View RGB}         & life statistics        \\
                                            &                                       & GPS, inventory, equipment        \\
                                            &                                       & location status~(biome, weather, \etc)         \\
\hline
\multirow{2}{*}{\textbf{{\mname}(ours)}}    & Ego-View RGB                          & life statistics                  \\
                                            & $3 \times 3 \times 3$ Voxels          & GPS, inventory, equipment        \\

\arrayrulecolor{black}
\bottomrule[1pt]
\end{tabular}
\label{tab:Observation Space}
\end{table*}

\subsection{Percipient}
\subsubsection{Data Collection}
\label{sup:Data Collection}


For data collection, we use Minedojo~\cite{fan2022minedojo} to obtain Minecraft snapshots which contain a wide array of details within the agent’s surroundings, including blocks, biomes, mobs and \etc. Following the environment creation, we enable our agent to perform a rotation on the spot, capturing snapshots from $12$ distinct perspectives spaced $30$ degrees apart. For each of these snapshots, we record the ground-truth information about the agent's surroundings by leveraging the data available in the MineDojo~\cite{fan2022minedojo} observation space such as Lidar. To ensure the exact correspondence between the ground-truth information and the image, the information corresponding to the Field of View region of the image is screened from the Lidar as the ground-truth information of the image.

To compile a comprehensive dataset encompassing various conditions and terrains in Minecraft, we implement a two-step data collection process: acquiring data related to different biomes and gathering data on different mobs. In the first step dedicated to gathering data on diverse biomes, we collect information from all $60$ biomes available in MineDojo~\cite{fan2022minedojo}. For each biome, we sample $20$ environments, resulting in a total of $7.2K$ images. 
In the second phase of gathering data for various mobs, our focus is on collecting images of $9$ commonly found mobs in the Minecraft world: zombies, skeletons, creepers, spiders, cows, chickens, sheep, pigs, and wolves. We specifically choose $30$ representative biomes from the available $60$ Minecraft biomes for this data batch. Among these $9$ types of mobs, the first four exclusively appear during the night, while the remaining five can be encountered both during the daytime and nighttime. For the mobs that appear in both periods, each mob type is generated across $30$ biomes, with $20$ environment samples ($10$ during the daytime and $10$ during the nighttime). This results in the creation of $36K$ images for these five mobs. As for the mobs exclusive to nighttime, they are generated in $30$ biomes, with $10$ nighttime environment samples per biome and $12$ images per environment sample, culminating in the generation of $7.2K$ images. 

The data obtained from both the first and second stages contribute to a comprehensive dataset totaling $50K$ images, and we prompt ChatGPT~\cite{chatgpt} to curate a list of instructions to obtain $500K$ image-text instruction-following data.

\subsubsection{MineLLM training details}

MineLLM combines the image visual encoder from MineCLIP~\cite{fan2022minedojo} and the large language models from Vicuna-13B-v1.5~\cite{chiang2023vicuna}. 
Images are processed by the frozen vision encoder, whose features are projected by a two-layer MLP named Alignment Net to the same feature space as the text embeddings of the applied LLM. 
Instructions are tokenized by SentencePiece tokenizer~\cite{kudo2018sentencepiece}, and then the vision and text tokens are concatenated to feed into the LLM model. 
To better align the feature space of visual image encoder from MineCLIP~\cite{fan2022minedojo} and large language model from Vicuna~\cite{chiang2023vicuna}, we collect $500K$ image-text instruction-following data on the MineDojo~\cite{fan2022minedojo} following the method detailed in \Cref{sup:Data Collection}, for the purpose of training MineLLM. 
Each training instance consists of an image $\mathcal{I}$ and a multi-turn conversation data $(\boldsymbol{x}_1, \boldsymbol{y}_1, \ldots, \boldsymbol{x}_n, \boldsymbol{y}_n)$, where $\boldsymbol{x}_i$ and $\boldsymbol{y}_i$ are the human’s instruction and the system’s response at the $i$-th turn. 
To train MineLLM efficiently, we add LoRA~\cite{hu2021lora} parameters to all projection layers of the self-attention layers in the LLM. 
Only the parameters of the Alignment Net and the LoRA~\cite{hu2021lora} module are optimized during training.  
Multimodal tokens are decoded by the LLM model and the corresponding LoRA~\cite{hu2021lora} parameters. 
The training objective of Percipient is defined as:
\begin{align}
    \mathcal{L}\left(\theta_{a}, \theta_{l}\right)=\prod_{i=1}^{n} p_{\theta}\left(\boldsymbol{y}_{i} \mid \boldsymbol{x}_{<i}, \boldsymbol{y}_{<i-1}, f\left(\mathcal{I}\right)\right),
\end{align}

where $\theta_{a}$ and $\theta_{l}$ correspond to the learnable parameters of the Alignment Net and LoRA~\cite{hu2021lora}. The $\mathcal{I}$ is the image representation produced by the visual encoder from MineCLIP~\cite{fan2022minedojo} and $\theta = \{ \theta_{a} , \theta_{l}, \theta_{m}, \theta_{v}\}$, where $\theta_{m}$ and $\theta_{v}$ are frozen parameters of MineCLIP~\cite{fan2022minedojo} and Vicuna-13B-v1.5~\cite{chiang2023vicuna}.
It is worth noting that during the training process, only system message responses denoted as $\boldsymbol{y}_{i}$, require loss computation.
Note that the loss is only computed from the part of system responses during training.
while training MineLLM, trainable parameters(\ie, $\theta_{a}$ from the Alignment Net and $\theta_{l}$ from LoRA~\cite{hu2021lora}) are optimized by Adam optimizer with a learning rate initialized to be $5e-4$, and scheduled using a linear decay scheduler.
The rank of LoRA~\cite{hu2021lora} modules is set to $32$. We train all parameters in a one-stage end-to-end fashion with $8$ A100 GPUs.
Each GPU process $2$ samples every iteration and the effective batch size is set to $128$ by gradient accumulation.
%
%
Input images are resized to be $224\times224$ and we use MineCLIP~\cite{fan2022minedojo} pre-trained ViT-B/16~\cite{dosovitskiy2020image} as visual encoder, the number of vision tokens are $196$ and length of text tokens after vision tokens are limited to $400$ in training.

\subsection{Memory}
%
Inspired by the Skill library of Voyager~\cite{wang2023voyager}, memory is utilized in two parts of {\mname} to perform Retrieval-Augmented Generation (RAG~\cite{lewis2020retrieval}). The Parser employs Knowledge Memory to decompose tasks into sub-objectives, while the Planner, when planning an action sequence for a specific sub-objective, may refer to similar action sequences provided by Performer Memory. The implementation details are similar to those of Voyager~\cite{wang2023voyager}.

\subsubsection{Knowledge Memory}
For Knowledge Memory, we actually adopt a vector database method (\eg, Chroma, FAISS, \etc) to store frequently used knowledge. This knowledge mainly comes from three sources: part of it is from the online wiki, another part is from the crafting recipes of items in MineDojo~\cite{fan2022minedojo}, and some are user tips from Reddit.
Specifically, we convert commonly used knowledge into corresponding text embeddings using OpenAI's text-embedding-ada-002~\cite{embeddingmodel} and store them in a vector database.
When decomposing sub-objectives requires the retrieval of relevant knowledge, we also convert the corresponding descriptions of these sub-objectives into corresponding text embeddings. We then perform a search match in the database and select the most similar piece of knowledge.
If the similarity score at this time is below $0.05$ (the lower the score, the more similar), it is directly taken as the result of the RAG~\cite{lewis2020retrieval}. Of course, there will also be cases where the similarity scores are all above $0.05$. This indicates that there is currently no such type of knowledge in the database. In this case, we manually supplement this type of knowledge and add it to the database as the result of the RAG~\cite{lewis2020retrieval}.

\subsubsection{Performer Memory}
For Performer Memory, we record the task description of each successful sub-objective and its corresponding successful action sequence.
Specifically, Performer Memory consists of two parts. One part is a vector database used to store the sub-objective task descriptions and their corresponding positions in the sub-objective sequence. The other part is a JSON file where the key is the position of the sub-objective in the sub-objective sequence, and the value corresponds to the sub-objective task description and its successful action sequence.
When we need to find similar action sequences, similar to Knowledge Memory, we convert the current sub-objective's task description into corresponding text embeddings and retrieve the $2$ closest matches from the vector library. We then extract the corresponding successful objective sequences from the JSON file using their positions in the sub-objective sequence.

\subsection{Observation Space}
In order to allow the system to more closely resemble an embodied agent rather than emulating a game player unlocking the tech tree, we significantly limited environmental information, endeavoring to enable the agent to perceive through Ego-View RGB images as much as possible.

Our Observation Space primarily consists of two components: one is the Perceptual Observation, and the other is the Status Observation. The Perceptual Observation includes Ego-View Minecraft-style RGB images and $3 \times 3 \times 3$ Voxels that the agent encounters. The Status Observation includes some associated auxiliary textual information(\eg, the current agent's life statistics, GPS location, inventory, and equipment information). Notably, to make the system more resemble an embodied agent, we have obscured a large amount of environmental information (\eg, the current biome, weather, and whether the sky is visible that human players can learn by pressing F3). This encourages the agent to perceive through the current RGB image rather than directly knowing a lot of the current environmental information.

To more clearly demonstrate our Observation Space, we list the differing Observation Spaces of related works in the table below, as shown in \Cref{tab:Observation Space}.

\subsection{Action Space}

\begin{table*}[htbp]
\caption{The Definition of the Action Space we use in MineDojo~\cite{fan2022minedojo} Simulator} 
\vspace{-3mm}
\centering
\small
\begin{tabular}{lllll}
\bottomrule[1pt]
 \multirow{2}{*}{Name}  & \multirow{2}{*}{Arguments} & \multirow{2}{*}{Description} & Corresponding   & Action Conditions Based on    \\
       &           &             & MineDojo~\cite{fan2022minedojo} Actions            & Environmental Information \\
\hline
  \multirow{2}{*}{Find} & \multirow{2}{*}{object}     & Travel across the present terrain &  forward, jump       & Halt only when the object is \\
                        &                             & in search of an object            &  move left and right & in Ego-View RGB image \\
\hline       
 \multirow{2}{*}{Move} & \multirow{2}{*}{object}     & Move to the target object until   &  forward, jump       & Halt only when the object is in the \\
                        &                                             & it is within striking distance    &  move left and right & surrounding $3 \times 3 \times 3$ Voxels \\
\hline
  \multirow{3}{*}{Craft}& object     & Craft a certain number of objects &  craft, attack       & Begins only once the environmental\\
                        & materials  & with materials in the inventory   &  use, place          & conditions required are met\\
                        & platform   & using the platform                &                      & \\
\hline
  \multirow{2}{*}{Mine} & object     & Harvest a single block using      &  \multirow{2}{*}{attack}              & Begins only once the environmental\\
                        & tool       & tool from surroundings            &                      & conditions required are met\\
\hline
  \multirow{2}{*}{Equip}& \multirow{2}{*}{object}     & Equip a given object from         &  \multirow{2}{*}{equip}               & Begins only once the environmental\\
                        &                             & the current inventory.            &                      & conditions required are met\\
\hline
  \multirow{2}{*}{Fight}& object     & Attack a nearby entity            &  \multirow{2}{*}{attack}              & Begins only once the environmental\\
                        & tool       & using the specified tool          &                      & conditions required are met\\
\hline 
  \multirow{2}{*}{Dig-Up}& \multirow{2}{*}{tool}      & Ascend directly by jumping        & \multirow{2}{*}{jump, place}          & Halt only when the agent\\
                         &                            & and placing  blocks               &                      & can see the sky\\
\hline
  \multirow{2}{*}{Dig-Down}& y-level & Descend using the specified tool  & \multirow{2}{*}{attack}               & Halt only when the agent \\
       & tool       & to dig your way through if necessary&                    & reach the specified y-level\\
\hline 
  \multirow{2}{*}{Use} & \multirow{2}{*}{object}    & Use the item held          & \multirow{2}{*}{use}                & \multirow{2}{*}{$\varnothing$}\\
                       &                            & in the main hand      &                      & \\
\hline 
  \multirow{2}{*}{Place} & \multirow{2}{*}{object}    & Place an inventory                & \multirow{2}{*}{place}                & Begins only once the environmental\\
        &           & item on the ground.               &                      & conditions required are met\\

\arrayrulecolor{black}
\bottomrule[1pt]
\end{tabular}
\label{tab:actions}
\end{table*}

The Performer module executes action sequences, which consist of actions falling within the action space outlined in Tab~\ref{tab:actions}. These actions are brief combinations formed by the MineDojo~\cite{fan2022minedojo} API, with frequent interactions with the environment occurring within each action. 

For example, the action of ``Find'' can be described as a directionless forward motion, initiating a jump when encountering obstacles. If the obstacle proves insurmountable, the action adapts by implementing a left or right turn, followed by the continuation of forward motion. This process involves minimal human intervention or design. 
During the execution of the ``Find'' action, there is a fixed frequency at which the current Ego-View RGB images are analyzed to ascertain whether the required object~(\eg, a block, a type of mob, \etc) has been in sight.

\section{Environment Setting}
\label{sup:Environment Setting}

\begin{table*}[]
\caption{Full \textit{Context-Dependent Tasks}. $16$ tasks are defined and divided into $4$ difficulty levels based on the minimum number of information types needed. \underline{Underlines} label the environmental information, reflecting the complexity varies at each level.}
\centering
\begin{tabular}{ccc}
\bottomrule[1pt]
Task Level& Task id  & Task description   \\
\hline
\multirow{4}{*}{Easy}
 &1-1  &Find a \underline{tree}~\raisebox{-0.3ex}{\includegraphics[width=0.3cm]{icon/tree.png}}      \\
 &1-2   &Find a \underline{grass}~\raisebox{-0.3ex}{\includegraphics[width=0.3cm]{icon/grass.png}}   \\
 &1-3  &Find a \underline{cow}~\raisebox{-0.3ex}{\includegraphics[width=0.3cm]{icon/cow.png}}    \\
 &1-4  &Find a \underline{pig}~\raisebox{-0.3ex}{\includegraphics[width=0.35cm]{icon/pig.png}}  \\
\hline
\multirow{4}{*}{Mid}
 &2-1  &Find a \underline{tree}~\raisebox{-0.3ex}{\includegraphics[width=0.3cm]{icon/tree.png}} in the \underline{forest}~\raisebox{-0.3ex}{\includegraphics[width=0.3cm]{icon/forest.png}}     \\
 &2-2  &Find a \underline{grass}~\raisebox{-0.3ex}{\includegraphics[width=0.3cm]{icon/grass.png}} near a \underline{pig}~\raisebox{-0.3ex}{\includegraphics[width=0.35cm]{icon/pig.png}}    \\
 &2-3  &Find a \underline{cow}~\raisebox{-0.3ex}{\includegraphics[width=0.3cm]{icon/cow.png}} in the \underline{desert}~\raisebox{-0.3ex}{\includegraphics[width=0.3cm]{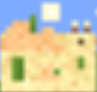}}    \\
 &2-4  &Find a \underline{pig}~\raisebox{-0.3ex}{\includegraphics[width=0.35cm]{icon/pig.png}} during the \underline{nighttime}~\raisebox{-0.3ex}{\includegraphics[width=0.3cm]{icon/moon.png}}   \\
\hline
\multirow{4}{*}{Hard}  
 &3-1  &Find a \underline{tree}~\raisebox{-0.3ex}{\includegraphics[width=0.3cm]{icon/tree.png}} in the \underline{forest}~\raisebox{-0.3ex}{\includegraphics[width=0.3cm]{icon/forest.png}} during the \underline{nighttime}~\raisebox{-0.3ex}{\includegraphics[width=0.3cm]{icon/moon.png}}      \\
 &3-2  &Find a \underline{grass}~\raisebox{-0.3ex}{\includegraphics[width=0.3cm]{icon/grass.png}} near a \underline{pig}~\raisebox{-0.3ex}{\includegraphics[width=0.35cm]{icon/pig.png}} in the \underline{plains}~\raisebox{-0.3ex}{\includegraphics[width=0.3cm]{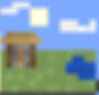}}   \\
 &3-3  &Find a \underline{cow}~\raisebox{-0.3ex}{\includegraphics[width=0.3cm]{icon/cow.png}} in the \underline{desert}~\raisebox{-0.3ex}{\includegraphics[width=0.3cm]{icon/desert.png}} during the \underline{daytime}~\raisebox{-0.3ex}{\includegraphics[width=0.3cm]{icon/sun.png}}   \\
 &3-4  &Find a \underline{pig}~\raisebox{-0.3ex}{\includegraphics[width=0.35cm]{icon/pig.png}} during the \underline{nighttime}~\raisebox{-0.3ex}{\includegraphics[width=0.3cm]{icon/moon.png}} in a \underline{rainy} day  \\
\hline
\multirow{5}{*}{Complex}
 &4-1  &Find a \underline{tree}~\raisebox{-0.3ex}{\includegraphics[width=0.3cm]{icon/tree.png}} in the \underline{forest}~\raisebox{-0.3ex}{\includegraphics[width=0.3cm]{icon/forest.png}} during the \underline{nighttime}~\raisebox{-0.3ex}{\includegraphics[width=0.3cm]{icon/moon.png}} in a \underline{sunny} day    \\
 &4-2   &Find a \underline{pig}~\raisebox{-0.3ex}{\includegraphics[width=0.35cm]{icon/pig.png}} near a \underline{grass}~\raisebox{-0.3ex}{\includegraphics[width=0.3cm]{icon/grass.png}} in the \underline{forest}~\raisebox{-0.3ex}{\includegraphics[width=0.3cm]{icon/forest.png}} during the \underline{daytime}~\raisebox{-0.3ex}{\includegraphics[width=0.3cm]{icon/sun.png}} \\ 
 &4-3   &Find a \underline{cow}~\raisebox{-0.3ex}{\includegraphics[width=0.3cm]{icon/cow.png}} near the \underline{water}~\raisebox{-0.3ex}{\includegraphics[width=0.3cm]{icon/water.png}} in the \underline{desert}~\raisebox{-0.3ex}{\includegraphics[width=0.3cm]{icon/desert.png}} during the \underline{daytime}~\raisebox{-0.3ex}{\includegraphics[width=0.3cm]{icon/sun.png}} in \underline{sunny} day  \\
 &\multirow{2}{*}{4-4}  &Find a \underline{pig}~\raisebox{-0.3ex}{\includegraphics[width=0.35cm]{icon/pig.png}} during the \underline{daytime}~\raisebox{-0.3ex}{\includegraphics[width=0.3cm]{icon/sun.png}} on the \underline{plains}~\raisebox{-0.3ex}{\includegraphics[width=0.3cm]{icon/plain.png}} with a \underline{grass}~\raisebox{-0.3ex}{\includegraphics[width=0.3cm]{icon/grass.png}} next to it,    \\
 &  &the weather is \underline{sunny} day and the \underline{brightness} is sufficient \\
   
\hline
\arrayrulecolor{black}
\bottomrule[1pt]
\end{tabular}
\label{tab:sup context task description}
\end{table*}

\begin{table*}[]
\caption{Details of \textit{Context-Dependent Tasks} Environment Information content.}
\centering
\begin{tabular}{ccccccccc}
\bottomrule[1pt]
Task Level &Task id&Num of Info.  & Object & Creature &Ecology &Time &Weather &Brightness \\
\hline
\multirow{4}{*}{Easy} &1-1 & 1   &\checkmark  & & & & &    \\
 &1-2 &1  &\checkmark   & & & & &   \\
 &1-3 &1 &  &\checkmark & & & &   \\
 &1-4 &1 &  &\checkmark & & & &  \\
\hline
\multirow{4}{*}{Medium} &2-1 &2  &\checkmark  & &\checkmark & & &   \\
 &2-2 &2 &\checkmark &\checkmark & & & &   \\
 &2-3 &2 &  &\checkmark &\checkmark & & &   \\
 &2-4 &2 & &\checkmark & &\checkmark & &  \\
\hline
\multirow{4}{*}{Hard} &3-1 &3  &\checkmark  & &\checkmark &\checkmark & &    \\
 &3-2 &3 &\checkmark &\checkmark &\checkmark & & &\\
 &3-3 &3 & &\checkmark &\checkmark &\checkmark & & \\
 &3-4 &3 & &\checkmark & &\checkmark &\checkmark &  \\
\hline
\multirow{4}{*}{Very Hard} &4-1 &4  &\checkmark &  &\checkmark &\checkmark &\checkmark &  \\
 &4-2 &4 &\checkmark  &\checkmark &\checkmark &\checkmark & &  \\
 &4-3 &5 &\checkmark &\checkmark &\checkmark &\checkmark &\checkmark & \\
 &4-4  &6 &\checkmark &\checkmark &\checkmark &\checkmark &\checkmark &\checkmark  \\
\hline
\arrayrulecolor{black}
\bottomrule[1pt]
\end{tabular}
\label{tab:sup context task info}
\end{table*}

Our Minecraft experimental environment is based on the MineDojo~\cite{fan2022minedojo} simulation platform, which provides a unified observation and action space to foster the development of intelligent agents capable of multitasking and continuous learning to adapt to new tasks and scenarios.

In our experiments, the \textit{position} at which the agent begins its game, as well as the \textit{seed} used to generate the environment, are both randomized. This introduces an element of unpredictability and variety into the experimental setup, ensuring that the agent will encounter a wide range of scenarios and challenges.
The agent is set to start in \textit{survival mode}, the most challenging and interactive mode available. Unlike creative or adventure modes, survival mode represents a test of the agent's ability to strategize, and make quick decisions. The agent is also confronted with the complication of \textit{hostile mob} generation.
The agent begins its game with an \textit{empty inventory}, meaning it must actively mine and craft the objects.
To simulate a real Embodied Agent, environmental factors(\eg, time, weather, \etc) change over time. At night, the agent does not have night vision, and the items in the inventory will be cleared upon death.

To better evaluate \textit{Context-Dependent Tasks} and \textit{Process-Dependent Tasks}, which are defined in detail in \Cref{sup:Task Details}, we select different environment settings in MineDojo~\cite{fan2022minedojo}. For \textit{Context-Dependent Tasks}, we uniformly adopt the environment in MineDojo~\cite{fan2022minedojo} with the creative ``task\_id'' of ``0''. For \textit{Process-Dependent Tasks}, we uniformly adopt the environment with the ``task\_id'' of ``harvest'', ``target\_names'' as ``diamond'', and ``spawn\_rate'' as ``1''. This is why obtaining redstone is more difficult than obtaining diamond, as described in \Cref{sup:Process-Dependent Tasks Main Results}.

\section{Task Details and Experiment Results}
\label{sup:Task Details and Experiment Results}

\begin{table*}[htbp]
\caption{Detailed Definition of \textit{Process-Dependent Tasks}. $25$ tasks are defined and divided into $5$ difficulty levels based on incrementally increasing reasoning steps. A higher difficulty level implies that the agent needs to engage in longer reasoning and planning with the environment.}
\centering
\begin{tabular}{cccccc}
\bottomrule[1pt]
Task Level&Task &reasoning step  & Object &Final recipe &Tools/Platforms     \\
\arrayrulecolor{black}\hline
\multirow{5}{*}{Basic level} 
&mine log  & 1  & ~\hspace{-0.3em}\raisebox{-0.3ex}{\includegraphics[width=1em,height=1em]{icon/wood.png}} &  - & -       \\
 &mine sand& 1 & ~\hspace{-0.3em}\raisebox{-0.3ex}{\includegraphics[width=1em,height=1em]{icon/sand.png}}  &  - & -        \\
 &craft planks & 2  & ~\hspace{-0.3em}\raisebox{-0.3ex}{\includegraphics[width=1em,height=1em]{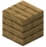}} &  1*~\hspace{-0.3em}\raisebox{-0.3ex}{\includegraphics[width=1em,height=1em]{icon/wood.png}} & -         \\
 &craft stick & 3 & \hspace{-0.3em}\raisebox{-0.3ex}{\includegraphics[width=1em,height=1em]{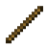}} &  2*~\hspace{-0.3em}\raisebox{-0.3ex}{\includegraphics[width=1em,height=1em]{icon/plank.png}} & -       \\
 &craft crafting table & 3 & ~\hspace{-0.3em}\raisebox{-0.3ex}{\includegraphics[width=1em,height=1em]{icon/crafting_table.png}} &  4*~\hspace{-0.3em}\raisebox{-0.3ex}{\includegraphics[width=1em,height=1em]{icon/plank.png}} & -         \\
\hline
\multirow{5}{*}{\makecell[c]{Wooden level}}  
 &craft bowl &4  &~\hspace{-0.3em}\raisebox{-0.3ex}{\includegraphics[width=1em,height=1em]{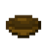}}  &3*~\hspace{-0.3em}\raisebox{-0.3ex}{\includegraphics[width=1em,height=1em]{icon/plank.png}} & \hspace{-0.3em}\raisebox{-0.3ex}{\includegraphics[width=1em,height=1em]{icon/crafting_table.png}}     \\
 &craft boat  &4  &~\hspace{-0.3em}\raisebox{-0.3ex}{\includegraphics[width=1em,height=1em]{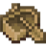}}  &5*~\hspace{-0.3em}\raisebox{-0.3ex}{\includegraphics[width=1em,height=1em]{icon/plank.png}} & \hspace{-0.3em}\raisebox{-0.3ex}{\includegraphics[width=1em,height=1em]{icon/crafting_table.png}}        \\
 &craft chest &4  &\raisebox{-0.3ex}{\includegraphics[width=0.3cm]{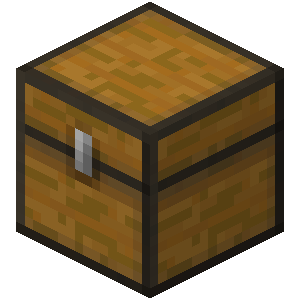}}  &8*~\hspace{-0.3em}\raisebox{-0.3ex}{\includegraphics[width=1em,height=1em]{icon/plank.png}} & \hspace{-0.3em}\raisebox{-0.3ex}{\includegraphics[width=1em,height=1em]{icon/crafting_table.png}}     \\
 &craft wooden sword &5  &\raisebox{-0.3ex}{\includegraphics[width=0.3cm]{icon/wooden_sword.png}}  &2*~\hspace{-0.3em}\raisebox{-0.3ex}{\includegraphics[width=1em,height=1em]{icon/plank.png}}+1* \hspace{-0.3em}\raisebox{-0.3ex}{\includegraphics[width=1em,height=1em]{icon/stick.png}} & \hspace{-0.3em}\raisebox{-0.3ex}{\includegraphics[width=1em,height=1em]{icon/crafting_table.png}}          \\
 &craft wooden pickaxe  &5 &  \hspace{-0.3em}\raisebox{-0.3ex}{\includegraphics[width=1em,height=1em]{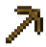}} &3*~\hspace{-0.3em}\raisebox{-0.3ex}{\includegraphics[width=1em,height=1em]{icon/plank.png}}+2* \hspace{-0.3em}\raisebox{-0.3ex}{\includegraphics[width=1em,height=1em]{icon/stick.png}} & \hspace{-0.3em}\raisebox{-0.3ex}{\includegraphics[width=1em,height=1em]{icon/crafting_table.png}}           \\
\hline
\multirow{5}{*}{\makecell[c]{Stone level}}  &mine cobblestone  &6  &\hspace{-0.3em}\raisebox{-0.3ex}{\includegraphics[width=1em,height=1em]{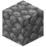}}    &-  & \hspace{-0.3em}\raisebox{-0.3ex}{\includegraphics[width=1em,height=1em]{icon/wooden_pickaxe.png}}     \\
 &craft furnace &7  &\hspace{-0.3em}\raisebox{-0.3ex}{\includegraphics[width=1em,height=1em]{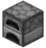}}    &8*~\hspace{-0.3em}\raisebox{-0.3ex}{\includegraphics[width=1em,height=1em]{icon/cobblestone.png}} & \hspace{-0.3em}\raisebox{-0.3ex}{\includegraphics[width=1em,height=1em]{icon/crafting_table.png}}      \\
 &craft stone pickaxe &7  &\hspace{-0.3em}\raisebox{-0.3ex}{\includegraphics[width=1em,height=1em]{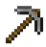}}   &3*~\hspace{-0.3em}\raisebox{-0.3ex}{\includegraphics[width=1em,height=1em]{icon/cobblestone.png}}+2* \hspace{-0.3em}\raisebox{-0.3ex}{\includegraphics[width=1em,height=1em]{icon/stick.png}} & \hspace{-0.3em}\raisebox{-0.3ex}{\includegraphics[width=1em,height=1em]{icon/crafting_table.png}}               \\
 &mine iron ore  &8  &  \hspace{-0.3em}\raisebox{-0.3ex}{\includegraphics[width=1em,height=1em]{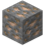}}  &-  &\hspace{-0.3em}\raisebox{-0.3ex}{\includegraphics[width=1em,height=1em]{icon/stone_pickaxe.png}}        \\
 &smelt glass  &9 &\hspace{-0.3em}\raisebox{-0.3ex}{\includegraphics[width=1em,height=1em]{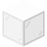}} &1*~\hspace{-0.3em}\raisebox{-0.3ex}{\includegraphics[width=1em,height=1em]{icon/sand.png}} & \hspace{-0.3em}\raisebox{-0.3ex}{\includegraphics[width=1em,height=1em]{icon/furnace.png}}         \\
\hline
\multirow{5}{*}{Iron level} &smelt iron ingot &10   &  \raisebox{-0.3ex}{\includegraphics[width=0.3cm]{icon/iron.png}}  &1*~\hspace{-0.3em}\raisebox{-0.3ex}{\includegraphics[width=1em,height=1em]{icon/iron_ore.png}} & \hspace{-0.3em}\raisebox{-0.3ex}{\includegraphics[width=1em,height=1em]{icon/furnace.png}}           \\
 &craft shield  &11  & \hspace{-0.3em}\raisebox{-0.3ex}{\includegraphics[width=0.3cm]{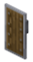}} & 1*\raisebox{-0.3ex}{\includegraphics[width=0.3cm]{icon/iron.png}}+6*~\hspace{-0.3em}\raisebox{-0.3ex}{\includegraphics[width=1em,height=1em]{icon/plank.png}}&\hspace{-0.3em}\raisebox{-0.3ex}{\includegraphics[width=1em,height=1em]{icon/crafting_table.png}}     \\
 &craft bucket &11  &  \hspace{-0.3em}\raisebox{-0.3ex}{\includegraphics[width=1em,height=1em]{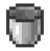}}  &3*\raisebox{-0.3ex}{\includegraphics[width=0.3cm]{icon/iron.png}} &\hspace{-0.3em}\raisebox{-0.3ex}{\includegraphics[width=1em,height=1em]{icon/crafting_table.png}}         \\
 &craft iron pickaxe  &11 & \hspace{-0.3em}\raisebox{-0.3ex}{\includegraphics[width=1em,height=1em]{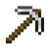}} &3*\raisebox{-0.3ex}{\includegraphics[width=0.3cm]{icon/iron.png}}+2* \hspace{-0.3em}\raisebox{-0.3ex}{\includegraphics[width=1em,height=1em]{icon/stick.png}} &\hspace{-0.3em}\raisebox{-0.3ex}{\includegraphics[width=1em,height=1em]{icon/crafting_table.png}}         \\
 &craft iron door  &11 & \raisebox{-0.3ex}{\includegraphics[width=0.3cm]{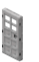}}&6*\raisebox{-0.3ex}{\includegraphics[width=0.3cm]{icon/iron.png}}  &\hspace{-0.3em}\raisebox{-0.3ex}{\includegraphics[width=1em,height=1em]{icon/crafting_table.png}}      \\
\hline
\multirow{5}{*}{Diamond level} &obtain diamond  &12  & \raisebox{-0.3ex}{\includegraphics[width=0.3cm]{icon/diamond.png}}   &-  & \hspace{-0.3em}\raisebox{-0.3ex}{\includegraphics[width=1em,height=1em]{icon/iron_pickaxe.png}}         \\
 &mind redstone &12   & \raisebox{-0.3ex}{\includegraphics[width=0.3cm]{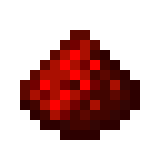}} &-  & \hspace{-0.3em}\raisebox{-0.3ex}{\includegraphics[width=1em,height=1em]{icon/iron_pickaxe.png}}          \\
 &craft compass &13   & \hspace{-0.1em}\raisebox{-0.3ex}{\includegraphics[width=1em,height=1em]{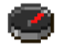}} &1*\raisebox{-0.3ex}{\includegraphics[width=0.3cm]{icon/redstone.png}}+4*\raisebox{-0.3ex}{\includegraphics[width=0.3cm]{icon/iron.png}} &\hspace{-0.3em}\raisebox{-0.3ex}{\includegraphics[width=1em,height=1em]{icon/crafting_table.png}}          \\
 &craft diamond pickaxe &13  & \hspace{-0.3em}\raisebox{-0.3ex}{\includegraphics[width=1em,height=1em]{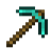}} &3*~\raisebox{-0.3ex}{\includegraphics[width=0.3cm]{icon/diamond.png}}+2* \hspace{-0.3em}\raisebox{-0.3ex}{\includegraphics[width=1em,height=1em]{icon/stick.png}} &\hspace{-0.3em}\raisebox{-0.3ex}{\includegraphics[width=1em,height=1em]{icon/crafting_table.png}}        \\
 
 &craft piston &13  & \hspace{-0.3em}\raisebox{-0.3ex}{\includegraphics[width=1em,height=1em]{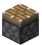}} &1*\raisebox{-0.3ex}{\includegraphics[width=0.3cm]{icon/redstone.png}}+1*\raisebox{-0.3ex}{\includegraphics[width=0.3cm]{icon/iron.png}}+4*~\hspace{-0.3em}\raisebox{-0.3ex}{\includegraphics[width=1em,height=1em]{icon/cobblestone.png}}+3*~\hspace{-0.1em}\raisebox{-0.3ex}{\includegraphics[width=1em,height=1em]{icon/plank.png}} &\hspace{-0.3em}\raisebox{-0.3ex}{\includegraphics[width=1em,height=1em]{icon/crafting_table.png}}      \\
\hline
\bottomrule[1pt]
\end{tabular}

\label{tab:Process-Dependent Tasks Definition}
\end{table*}

\subsection{Task Details}
\label{sup:Task Details}

\subsubsection{Context-Dependent Tasks}
\textit{Context-Dependent Tasks} primarily study how Active Perception enables the agent to better perceive low-level context information in the environment.
We first establish $6$ aspects of environmental information derived from the Minecraft game environment:  \textsl{[Object, Mob, Ecology, Time, Weather, Brightness]}. Each aspect has multiple options. For example, pigs~\raisebox{-0.3ex}{\includegraphics[width=0.35cm]{icon/pig.png}}, cows~\raisebox{-0.3ex}{\includegraphics[width=0.3cm]{icon/cow.png}}, and sheep~\raisebox{-0.3ex}{\includegraphics[width=0.3cm]{icon/sheep.png}} are all elements belonging to Mob. Based on this, we define $16$ tasks and organize their difficulty into $4$ levels by taking into account the number of information elements that require perception, as is shown in Tab.~\ref{tab:sup context task description}. Easy tasks necessitate the perception of only one element, Mid tasks include $2$ perception elements, Hard tasks contain $3$ elements, whereas Complex tasks involve the perception of $4$ to $6$ elements. Each task at the same level has different environment information content, the amount of environment information contained in each task, and the corresponding specific environment information is shown in Tab.~\ref{tab:sup context task info}. Finally, we rigorously assess {\mname}'s proficiency in environmental context perception across these $16$ tasks. 

As the main paper states, our initial environmental details are predetermined (\eg, biomes) in order to reduce the agent's exploration time, otherwise, the agent may fail to find the corresponding scenario within the time limit. We defined ten initial biome, each of which used random seeds to generate five different environments to test each task, so each task was tested in $50$ different scenarios and the success rate was calculated to verify {\mname}'s generalization ability. In order to align as much as possible with the experimental Settings of other methods, we did not modify the terrain to simplify the task.

\subsubsection{Process-Dependent Tasks}

\begin{figure*}
\centering
\includegraphics[width=1\linewidth]{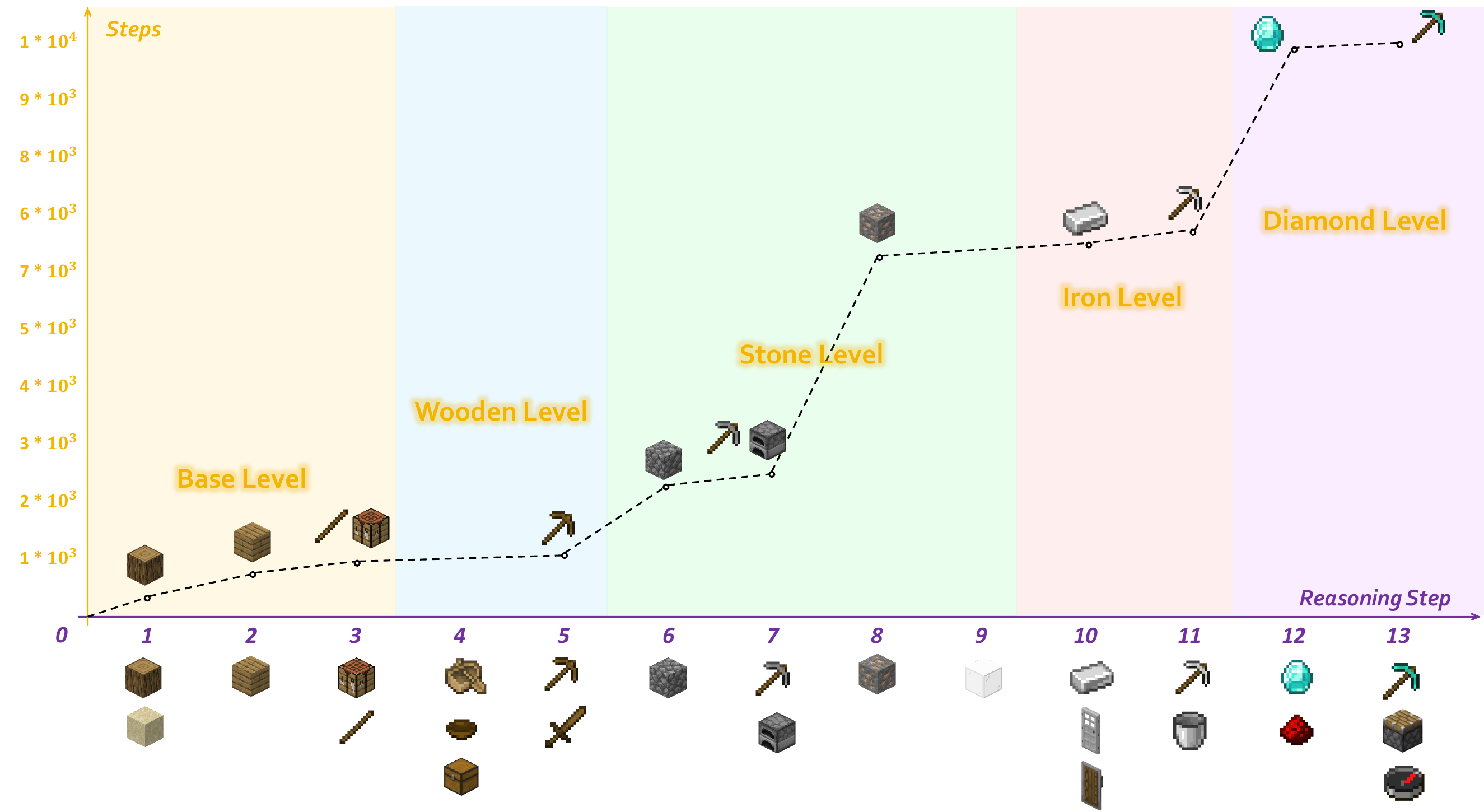}
   \caption{The game-playing steps corresponding to the acquisition of different milestone objects by the agent during the completion of the \textit{craft diamond pickaxe} challenge. The varying background colors denote the level of the \textit{Process-Dependent Tasks} in which the milestone objects are located.}
\label{fig:step}
\end{figure*}

\textit{Process-Dependent Tasks} primarily investigate situation-aware planning and embodied action execution, incorporating contributions from Active Perception and other modules that continuously perceive the environment and dynamically adjust their actions to accomplish long-horizon tasks. 
In \Cref{tab:Process-Dependent Tasks Definition}, we list the names of all tasks in \textit{Process-Dependent Tasks}, their reasoning steps, object icons, the final recipe, and the required tools/platforms.
The reasoning step refers to the number of sub-objectives that need to be completed in order to finish the entire task. Given that the agent's environment information(\eg, biome, weather, \etc) is randomly initialized, there may be execution errors requiring replanning, thus potentially necessitating the completion of additional sub-objectives, which means more reasoning steps may be required.
We consider only the most basic scenarios and select $25$ tasks based on the required reasoning steps in increasing order. These tasks are then divided into $5$ difficulty levels.

For evaluation, we consider an Agent's accidental death in the game (\eg, being burned by lava, killed by a hostile mob, \etc) as a failure, as well as not achieving the objective within the time limit (\eg, exceeding the $10$ minute game limit, or API request timeout, \etc). We conduct $30$ games of \textit{Process-Dependent Tasks} and took the average success rate as the final reported performance.

\subsection{Success Rates of All Tasks}
\subsubsection{Context-Dependent Tasks}

We report the success rates of different methods and perception strategies for all tasks comprehensively and in detail in \Cref{tab:sup main results for Context-Dependent Tasks}, including ours, GPT-4V~\cite{gpt4v}, and LLaVA-1.5~\cite{liu2023improved}, using both Active Perception strategy and Fine-Grained Global Perception strategy. This table also presents the detailed results of the ``Main Results'' section under ``\textit{Context-Dependent Tasks}'' in the main text.

\begin{table*}[htbp]
\centering
\caption{Detailed Performance on \textit{Context-Dependent Tasks}. Method$_A$ means the method uses the Active Perception strategy, and Method$_G$ means the method uses the Fine-Grained Global Perception strategy. The parts with a gray background in the table represent the average success rate for the current level.}
\begin{tabular}{cc|cccccc}
\bottomrule[1pt]
\multirow{2}{*}{Task Level} & \multirow{2}{*}{Task id} & \multicolumn{6}{c}{Success rate(\%)} \\

& & {\mname}$_A$ & {\mname}$_G$ & GPT-4V$_A$~\cite{gpt4v} & GPT-4V$_G$~\cite{gpt4v} &LLaVA-1.5$_A$~\cite{liu2023improved} &LLaVA-1.5$_G$~\cite{liu2023improved}\\
\hline
\multirow{5}{*}{\makecell[c]{Easy}} 
 &1-1             & 98.0 & 94.0 & 100.0   &100.0 &88.0&56.0\\
 &1-2             & 100.0 & 92.0 & 100.0  &100.0 &68.0&44.0\\
 &1-3             & 98.0 & 88.0 & 100.0   &96.0  &76.0&42.0\\
 &1-4             & 98.0 & 86.0 & 100.0   &94.0 &58.0&48.0\\
 &\cellcolor{gray!15}{Average}         & \cellcolor{gray!15}{98.5} & \cellcolor{gray!15}{90.0} & \cellcolor{gray!15}{100.0}   &\cellcolor{gray!15}{97.5}  &\cellcolor{gray!15}{72.5}&\cellcolor{gray!15}{47.5} \\
\hline
\multirow{5}{*}{\makecell[c]{Mid}}  
 &2-1             & 98.0 & 90.0 & 98.0  &82.0  &56.0&28.0\\
 &2-2             & 96.0 & 82.0 & 90.0  &86.0 &52.0&14.0\\
 &2-3             & 92.0 & 88.0 & 94.0  &88.0  &44.0&22.0\\
 &2-4             & 92.0 & 84.0 & 96.0  &84.0  &48.0&26.0\\
 &\cellcolor{gray!15}{Average}         & \cellcolor{gray!15}{94.5} & \cellcolor{gray!15}{86.0} & \cellcolor{gray!15}{94.5} &\cellcolor{gray!15}{85.0}  &\cellcolor{gray!15}{50.0}&\cellcolor{gray!15}{22.5}\\
\hline
\multirow{5}{*}{\makecell[c]{Hard}}  
 &3-1            & 94.0 & 80.0  & 96.0   &80.0  &12.0&8.0\\
 &3-2            & 98.0 & 78.0  & 92.0   &74.0  &8.0 &0.0\\
 &3-3            & 90.0 & 76.0  & 90.0   &74.0  &10.0&6.0\\
 &3-4            & 90.0 & 76.0  & 92.0   &72.0  &14.0&6.0\\
&\cellcolor{gray!15}{Average}         & \cellcolor{gray!15}{93.0} &\cellcolor{gray!15}{77.5}  & \cellcolor{gray!15}{92.5}   &\cellcolor{gray!15}{75.0}  &\cellcolor{gray!15}{11.0}&\cellcolor{gray!15}{5.0}\\
\hline
\multirow{5}{*}{Complex} 
 &4-1             & 92.0 & 74.0 & 90.0&64.0 &0.0&0.0 \\
 &4-2             & 92.0 & 70.0 & 88.0&60.0 &0.0&0.0\\
 &4-3             & 86.0 & 64.0 & 84.0&58.0 &0.0&0.0\\
 &4-4             & 94.0 & 62.0 & 88.0&58.0 &0.0&0.0\\
 &\cellcolor{gray!15}{Average}         & \cellcolor{gray!15}{91.0} & \cellcolor{gray!15}{67.5} & \cellcolor{gray!15}{87.5}&\cellcolor{gray!15}{60.0} &\cellcolor{gray!15}{0.0}&\cellcolor{gray!15}{0.0}\\
\bottomrule[1pt]

\end{tabular}
\label{tab:sup main results for Context-Dependent Tasks}
\end{table*}

\begin{table*}[htbp]
\centering
\caption{Detailed Ablation on \textit{Context-Dependent Tasks}. The parts with a gray background in the table represent the average success rate for the current level.}
\begin{tabular}{cc|ccc}
\bottomrule[1pt]
\multirow{2}{*}{Task Level} & \multirow{2}{*}{Task id} & \multicolumn{3}{c}{Success rate(\%)} \\

& & MineLLM+MineCLIP~\cite{fan2022minedojo} & MineLLM+CLIP~\cite{radford2021learning}  &LLaVA-1.5~\cite{liu2023improved}+CLIP~\cite{radford2021learning}\\
\hline
\multirow{5}{*}{\makecell[c]{Easy}} 
 &1-1             & 98.0 & 98.0   &88.0\\
 &1-2             & 100.0 & 94.0  &68.0\\
 &1-3             & 98.0 & 96.0   &76.0\\
 &1-4             & 98.0 & 92.0   &58.0\\
 &\cellcolor{gray!15}{Average}         & \cellcolor{gray!15}{98.5} & \cellcolor{gray!15}{95.0}   & \cellcolor{gray!15}{72.5} \\
\hline
\multirow{5}{*}{\makecell[c]{Mid}}  
 &2-1             & 98.0 & 94.0   &56.0\\
 &2-2             & 96.0 & 88.0   &52.0\\
 &2-3             & 92.0 & 88.0   &44.0\\
 &2-4             & 92.0 & 90.0   &48.0\\
 &\cellcolor{gray!15}{Average}         & \cellcolor{gray!15}{94.5} & \cellcolor{gray!15}{90.0}  &\cellcolor{gray!15}{50.0}\\
\hline
\multirow{5}{*}{\makecell[c]{Hard}}  
 &3-1            & 94.0 & 90.0    &12.0\\
 &3-2            & 98.0 & 90.0    &8.0 \\
 &3-3            & 90.0 & 84.0    &10.0\\
 &3-4            & 90.0 & 84.0    &14.0\\
&\cellcolor{gray!15}{Average}         & \cellcolor{gray!15}{93.0} & \cellcolor{gray!15}{87.0}    &\cellcolor{gray!15}{11.0}\\
\hline
\multirow{5}{*}{Complex} 
 &4-1             & 92.0 & 82.0  &0.0 \\
 &4-2             & 92.0 & 84.0  &0.0\\
 &4-3             & 86.0 & 78.0  &0.0\\
 &4-4             & 90.0 & 76.0  &0.0\\
 &\cellcolor{gray!15}{Average}         & \cellcolor{gray!15}{91.0} & \cellcolor{gray!15}{80.0}  &\cellcolor{gray!15}{0.0}\\
\bottomrule[1pt]

\end{tabular}
\label{tab:sup ab results for Context-Dependent Tasks}
\end{table*}



\subsubsection{Process-Dependent Tasks}
\label{sup:Process-Dependent Tasks Main Results}

We report the success rates of different methods for all tasks comprehensively and in detail in \Cref{tab:main results for Process-Dependent Tasks in sup}, including ours, non-situation-aware planning, and non-embodied action execution. This table also presents the detailed results of the ``Main Results'' section under ``\textit{Process-Dependent Tasks}'' in the main text. The parts with a gray background in the table represent the average success rate for the current level.

To better demonstrate the practical performance of {\mname} in \textit{Process-Dependent Tasks}, we select \textit{craft diamond pickaxe} with a reasoning step of $13$ as the challenge. \Cref{fig:step} depicts the game-playing steps corresponding to each milestone object~(\eg, log~\raisebox{-0.3ex}{\includegraphics[width=1em,height=1em]{icon/wood.png}}, plank~\raisebox{-0.3ex}{\includegraphics[width=1em,height=1em]{icon/plank.png}}, stick~\raisebox{-0.3ex}{\includegraphics[width=1em,height=1em]{icon/stick.png}}, \etc) obtained by the agent.

\begin{table*}[htbp]
\centering
\caption{Detailed Performance on \textit{Process-Dependent Tasks}. We compare the success rate when interacting or not interacting with the environment during the planning or execution. The parts with a gray background in the table represent the average success rate for the current level.}
\begin{tabular}{cc|ccc}
\bottomrule[1pt]
\multirow{2}{*}{Task Level} & \multirow{2}{*}{Object} & \multicolumn{3}{c}{Success rate(\%)} \\

& & {\mname}(Ours) & non-situation-aware planning & non-embodied action execution\\
\hline
\multirow{6}{*}{\makecell[c]{Basic level}} 
 &log             & 96.67 & 93.33 & 0.00  \\
 &sand            & 96.67 & 93.33 & 0.00  \\
 &planks          & 96.67 & 93.33 & 0.00   \\
 &stick           & 96.67 & 90.00 & 0.00    \\
 &crafting table  & 93.33 & 90.00 & 0.00    \\
 &\cellcolor{gray!15}{Average} & \cellcolor{gray!15}{96.00} & \cellcolor{gray!15}{92.00} & \cellcolor{gray!15}{0.00} \\
\hline
\multirow{6}{*}{\makecell[c]{Wooden level}}  
 &bowl            & 93.33 & 90.00 & 0.00\\
 &boat            & 93.33 & 90.00 & 0.00 \\
 &chest           & 90.00 & 90.00 & 0.00\\
 &wooden sword    & 86.67 & 80.00 & 0.00\\
 &wooden pickaxe  & 80.00 & 80.00 & 0.00 \\
 &\cellcolor{gray!15}{Average} & \cellcolor{gray!15}{88.67} & \cellcolor{gray!15}{86.00} & \cellcolor{gray!15}{0.00} \\
\hline
\multirow{6}{*}{\makecell[c]{Stone level}}  
 &cobblestone     & 80.00 & 73.33  & 0.00 \\
 &furnace         & 80.00 & 73.33  & 0.00  \\
 &stone pickaxe   & 80.00 & 70.00  & 0.00 \\
 &iron ore        & 60.00 & 50.00  & 0.00\\
 &glass           & 80.00 & 76.67  & 0.00 \\
&\cellcolor{gray!15}{Average} & \cellcolor{gray!15}{76.00} & \cellcolor{gray!15}{68.67} & \cellcolor{gray!15}{0.00}\\
\hline
\multirow{6}{*}{Iron level} 
 &iron ingot       & 56.67 & 50.00 & 0.00  \\
 &shield          & 56.67 & 50.00 & 0.00\\
 &bucket          & 53.33 & 43.33 & 0.00\\
 &iron pickaxe    & 50.00 & 40.00 & 0.00\\
 &iron door       & 43.33 & 43.33 & 0.00\\
 &\cellcolor{gray!15}{Average} & \cellcolor{gray!15}{52.00} & \cellcolor{gray!15}{45.33} & \cellcolor{gray!15}{0.00}\\
\hline
\multirow{6}{*}{Diamond level} 
 &diamond ore     & 30.00 & 20.00 & 0.00 \\
 &mind redstone   & 20.00 & 16.67 & 0.00\\
 &compass         & 16.67 & 10.00 & 0.00\\
 &diamond pickaxe & 23.33 & 10.00 & 0.00  \\
 &piston          & 20.00 & 13.33 & 0.00\\
 &\cellcolor{gray!15}{Average} & \cellcolor{gray!15}{22.00} & \cellcolor{gray!15}{14.00} & \cellcolor{gray!15}{0.00} \\
\bottomrule[1pt]

\end{tabular}
\label{tab:main results for Process-Dependent Tasks in sup}
\end{table*}

\subsection{Ablation Study}
\subsubsection{Context-Dependent Tasks}

We conduct ablation studies on the multi-modal large language model (MLLM) part within \textit{Context-Dependent Tasks} in \ref{tab:sup ab results for Context-Dependent Tasks}, comparing the performance outcomes of different MLLMs and different pre-trained visual encoders in the percipient.

\subsubsection{Process-Dependent Tasks}
In this section, we present detailed results from our ablation experiments. 
\Cref{tab:ablation for all components in sup} shows the performance of the agent in {\mname} after the removal of various modules.
\Cref{tab:ablation for planner in sup} demonstrates the impact on the results when the Planner is replaced by large language models with inconsistent reasoning capabilities, including open-source models like LLaMA2-70B-Chat~\cite{touvron2023llama2} and Vicuna-13B-v1.5-16k~\cite{chiang2023vicuna}. 
\Cref{tab:ablation for memory in sup} further explores the contribution of the Memory components to the agent's performance, including Knowledge Memory and Performer Memory. 
\Cref{tab:ablation for all patroller in sup} investigates the robustness gain brought by the check part of the Patroller under \textit{``Random Drop''} conditions.
As seen from the results in \Cref{tab:ablation for all components in sup}, the agent's success rate in completing \textit{Process-Dependent Tasks} significantly decreases after the removal of any modules, with the success rate at the Diamond level~\raisebox{-0.3ex}{\includegraphics[width=0.3cm]{icon/diamond.png}}  falling to $0.00\%$ for all except when the Patroller is removed. The Percipient mainly provides the agent with visual input, the Memory primarily provides the agent with relevant knowledge, the Parser simplifies the difficulty of online task decomposition for the agent, and the Patroller ensures that each action is sufficiently checked for successful execution.
\Cref{tab:ablation for planner in sup} presents detailed results from the Planner ablation experiments in the ``Ablation Study'' section of the main text. From this, we can discern that LLMs with stronger reasoning capabilities demonstrate better understanding when faced with a wide variety of text information inputs, thereby facilitating more effective planning. The poor performance of open-source large models like LLaMA2-70B-Chat~\cite{touvron2023llama2} Vicuna-13B-v1.5-16k~\cite{chiang2023vicuna} is due to their inadequate ability to process long and diverse types of text information. This inadequacy is evident at the Wooden level~\raisebox{-0.3ex}{\includegraphics[width=0.3cm]{icon/wooden_sword.png}}, where the success rate has already plummeted to $0.00\%$.
As can be seen from the results in \Cref{tab:ablation for memory in sup}, both types of Memory can enhance the agent's actions, particularly the Knowledge Memory. Without the Knowledge Memory, the agent fails to mine iron due to its inability to recognize where iron ore is more likely to be located. Consequently, the success rates for both Iron~\raisebox{-0.3ex}{\includegraphics[width=0.3cm]{icon/iron.png}} and Diamond levels~\raisebox{-0.3ex}{\includegraphics[width=0.3cm]{icon/diamond.png}} are $0.00\%$. The Knowledge Memory can help the agent more easily understand the acquisition methods of some items, while the Performer Memory can provide similar scenarios for the agent to reference, thereby easing the pressure in the planning process.
\Cref{tab:ablation for all patroller in sup} primarily studies the robustness brought about by the check part of the Patroller. \textit{``Random Drop''} is a specific setting that forces the Agent into execution errors. More specifically, when the agent successfully completes tasks with the reasoning step greater than $4$, it will randomly discard one item from either log~\raisebox{-0.3ex}{\includegraphics[width=1em,height=1em]{icon/wood.png}}, planks~\raisebox{-0.3ex}{\includegraphics[width=1em,height=1em]{icon/plank.png}}, or stick~\raisebox{-0.3ex}{\includegraphics[width=1em,height=1em]{icon/stick.png}} present in its inventory.
This situation can lead the agent to commit execution errors due to insufficient material, specifically when it is completing sub-objectives of higher reasoning steps that require logs~\raisebox{-0.3ex}{\includegraphics[width=1em,height=1em]{icon/wood.png}}, planks~\raisebox{-0.3ex}{\includegraphics[width=1em,height=1em]{icon/plank.png}}, or sticks~\raisebox{-0.3ex}{\includegraphics[width=1em,height=1em]{icon/stick.png}} as materials. The check part of the Patroller can detect the cause of these errors during execution and use it as feedback for re-planning. With the \textit{``Random Drop''} enabled and the check part of the Patroller disabled, the agent even struggles to complete tasks at the stone level~\raisebox{-0.3ex}{\includegraphics[width=0.3cm]{icon/stone.png}} are $0.00\%$ effectively.

\section{Different Strategy of Active Perception}
\label{sup:Different Strategy of Active Perception}
In order to improve the quality of the Active Perception Query generated by Patroller, we use Chain-of-Thought(COT)\cite{wei2022chain} to design a process of multiple rounds of query generation, Patroller can generate the next most important problem based on the current problem and task description, until the agent judges that all problems have been produced. We conduct experiences to compare Single-round Generation and Multi-round Generation in Tab.~\ref{tab:multi-turn exp}, We can observe that Multi-round Generation using COT\cite{wei2022chain} generates better corresponding environment information query and thus have a higher success rate on the \textit{Context-Dependent Tasks}.

\begin{table}[]
\caption{Performance on Active Perception Query Generation with different Round Strategy. S means Single-round Generation and M means Multi-round Generation.}
\centering
\scriptsize
\begin{tabular}{lc|ccccc}
\bottomrule[1pt]
\multirow{2}{*}{planner} &\multirow{2}{*}{Strategy} &  \multicolumn{4}{c}{Average Generation Rate(\%)} \\
                        &  &   Easy & Mid & Hard & Complex  \\
\hline
\multirow{2}{*}{Vicuna-13B-v1.5~\cite{chiang2023vicuna}}  &S  &  100 & 95 & 75 & 45  \\  
    & M & 100 & 100 & 95 & 80  \\  
\hline
\multirow{2}{*}{GPT-3.5-turbo~\cite{chatgpt}} &S &  100 & 100 & 85 & 70  \\  
    & M & 100 & 100 & 100 & 100  \\  
\arrayrulecolor{black}
\bottomrule[1pt]
\end{tabular}
\label{tab:multi-turn exp}
\end{table}

\section{Applications}
\label{sup:Applications}

\subsection{Obtain Diamond Pickaxe}
We demonstrate a case of the popular \textit{Process-Dependent Tasks} ``\textit{craft diamond pickaxe}~\raisebox{-0.3ex}{\includegraphics[width=0.3cm]{icon/diamond_pickaxe.png}}'' challenge in Video 1.

\subsection{Discovery}
We demonstrate a complex level \textit{Context-Dependent Tasks} ``\textit{Find a pig~\raisebox{-0.3ex}{\includegraphics[width=0.35cm]{icon/pig.png}} on the plains~~\raisebox{-0.3ex}{\includegraphics[width=0.3cm]{icon/plain.png}} with grass~\raisebox{-0.3ex}{\includegraphics[width=0.3cm]{icon/grass.png}} and water~\raisebox{-0.3ex}{\includegraphics[width=0.3cm]{icon/water.png}} next to it during a sunny day with sufficient brightness}'' in Video 2.

\subsection{Open-Ended Tasks}
We demonstrate a \textit{Open-Ended Tasks} ``\textit{Dig a block of sand~\raisebox{-0.3ex}{\includegraphics[width=0.3cm]{icon/sand.png}} under the water~\raisebox{-0.3ex}{\includegraphics[width=0.3cm]{icon/water.png}} with a wooden shovel~\raisebox{-0.3ex}{\includegraphics[width=0.3cm]{icon/wooden_shovel.png}} during the daytime~\raisebox{-0.3ex}{\includegraphics[width=0.3cm]{icon/sun.png}} on a sunny day}'' in Video 3.

\section{Interactions in {\mname}}
\label{sup:Interactions}

Here we illustrate the interactions between the internal modules of {\mname} during Active Perception and Re-planning, presented in the form of dialogue text.

\subsection{Active Perception}
\label{sup:Interactions of Active Perception}

\begin{figure*}
\centering
\includegraphics[width=1\linewidth]{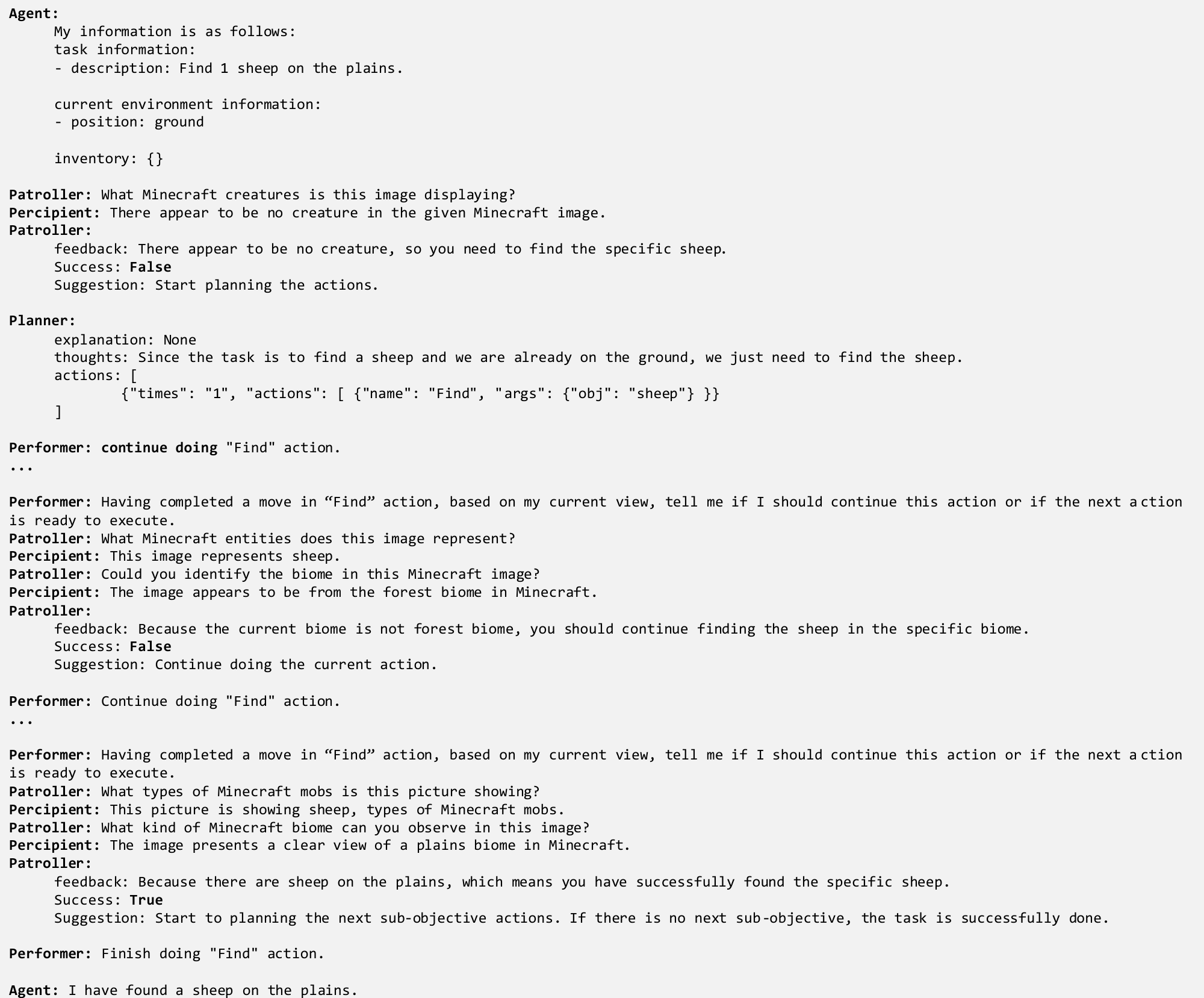}
   \caption{Dialogue of task ``\textit{Find 1 sheep~\raisebox{-0.3ex}{\includegraphics[width=0.3cm]{icon/sheep.png}} on the plains~\raisebox{-0.3ex}{\includegraphics[width=0.3cm]{icon/plain.png}}}'' }
\label{fig:Interactions of Active Perception}
\end{figure*}

\begin{figure*}
\centering
\includegraphics[width=1\linewidth]{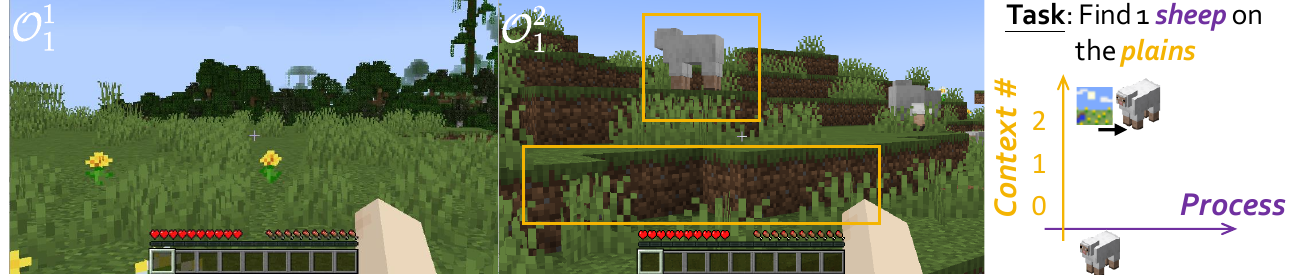}
   \caption{The corresponding screenshots for the dialogue of task ``\textit{Find 1 sheep~\raisebox{-0.3ex}{\includegraphics[width=0.3cm]{icon/sheep.png}} on the plains~\raisebox{-0.3ex}{\includegraphics[width=0.3cm]{icon/plain.png}}}''}
\label{fig:Screenshots of Active Perception}
\end{figure*}

In this part, we demonstrate the communication process among situation-aware planning, embodied action execution, and active perception scheme when facing the task of ``\textit{Find 1 sheep~\raisebox{-0.3ex}{\includegraphics[width=0.3cm]{icon/sheep.png}}} on the plains~\raisebox{-0.3ex}{\includegraphics[width=0.3cm]{icon/plain.png}}'', as shown in \Cref{fig:Interactions of Active Perception}. The corresponding screenshots are illustrated in \Cref{fig:Screenshots of Active Perception}.

\subsection{Re-planning}
\label{sup:Interactions of Re-planning}
\begin{figure*}
\centering
\includegraphics[width=1\linewidth]{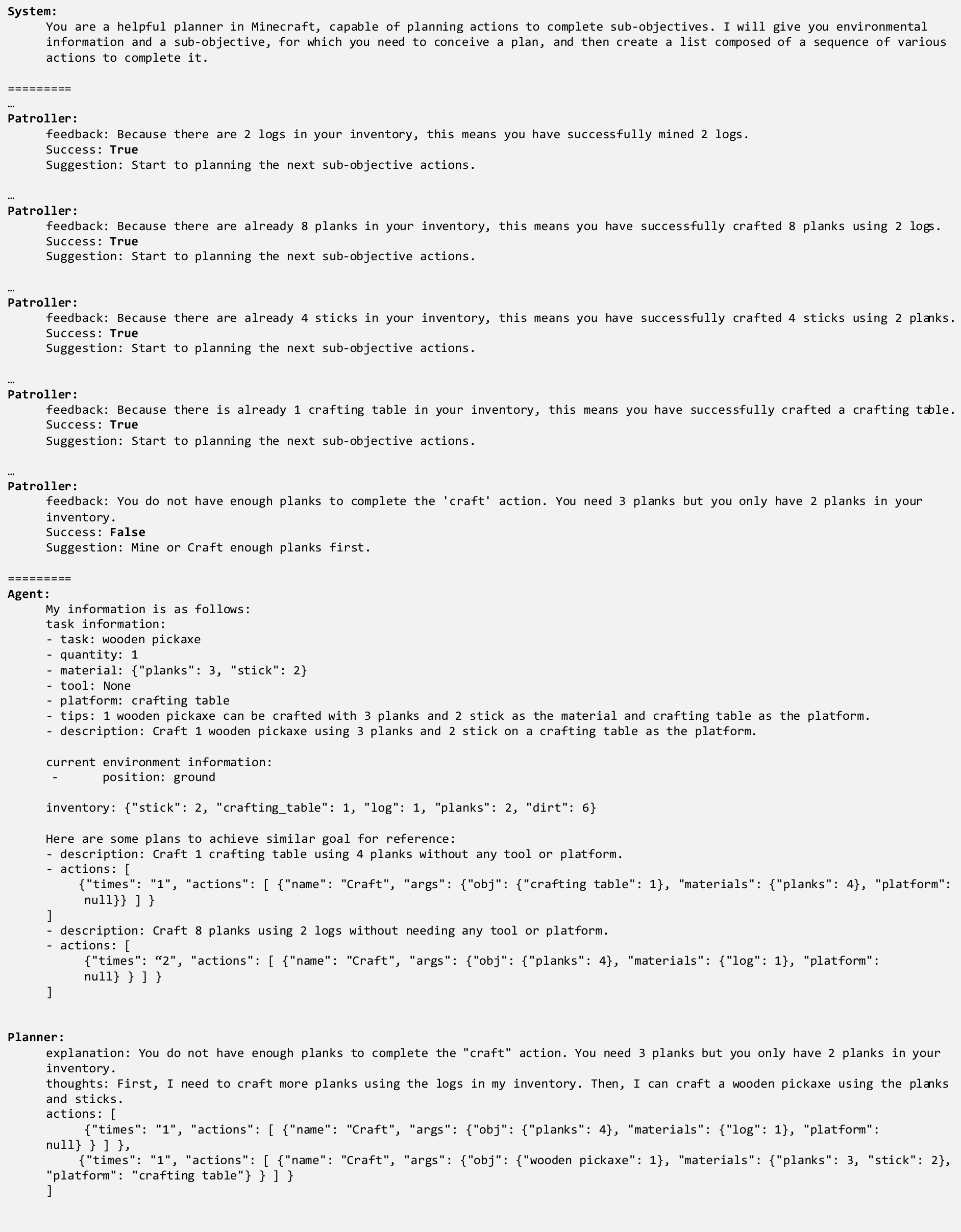}
   \caption{Dialogue of task ``\textit{craft wooden pickaxe}~\hspace{-0.3em}\raisebox{-0.3ex}{\includegraphics[width=1em,height=1em]{icon/wooden_pickaxe.png}}'' while re-planning}
\label{fig:Interactions of Re-planning}
\end{figure*}

\begin{figure*}
\centering
\includegraphics[width=1\linewidth]{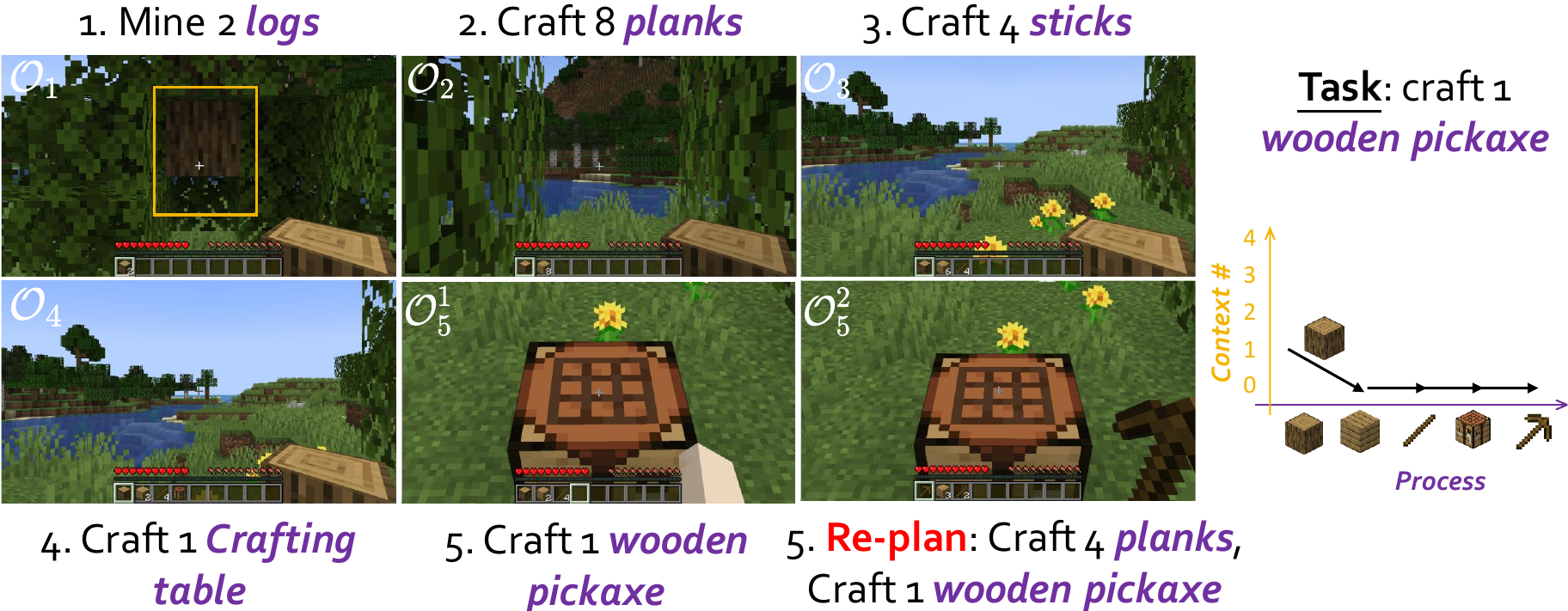}
   \caption{The corresponding screenshots for the dialogue of task ``\textit{Find 1 sheep~\raisebox{-0.3ex}{\includegraphics[width=0.3cm]{icon/sheep.png}} on the plains~\raisebox{-0.3ex}{\includegraphics[width=0.3cm]{icon/plain.png}}}'' while re-planning}
\label{fig:Screenshots of Re-planning}
\end{figure*}

In this part, we depict the situation when facing the task of ``\textit{craft wooden pickaxe}~\hspace{-0.3em}\raisebox{-0.3ex}{\includegraphics[width=1em,height=1em]{icon/wooden_pickaxe.png}}'' with a shortfall of 1 plank~\raisebox{-0.3ex}{\includegraphics[width=1em,height=1em]{icon/plank.png}}. In this case, the Patroller identifies the cause of the execution error and instructs the Planner to re-plan, as shown in \Cref{fig:Interactions of Re-planning}. The corresponding screenshots are illustrated in \Cref{fig:Screenshots of Re-planning}.



\begin{table*}[htbp]
\centering
\caption{Success rates on different modules within \textit{Process-Dependent Task}. The parts with a gray background in the table represent the average success rate for the current level.}
\begin{tabular}{cc|ccccc}
\bottomrule[1pt]

\multirow{2}{*}{Task Level} & \multirow{2}{*}{Object} & \multicolumn{5}{c}{Success rate(\%)} \\

& & {\mname}(Ours) & w/o Percipient & w/o Memory & w/o Parser & w/o Patroller \\
\hline
\multirow{6}{*}{\makecell[c]{Basic level}} 
 &log             & 96.67 & 0.00 & 90.00 & 96.67 & 86.67      \\
 &sand            & 96.67 & 0.00 & 90.00 & 96.67 & 73.33    \\
 &planks          & 96.67 & 0.00 & 80.00 & 96.67 & 83.33      \\
 &stick           & 96.67 & 0.00 & 76.67 & 96.67 & 73.33    \\
 &crafting table  & 93.33 & 0.00 & 76.67 & 90.00 & 73.33      \\
 &\cellcolor{gray!15}{Average} & \cellcolor{gray!15}{96.00} & \cellcolor{gray!15}{0.00} &\cellcolor{gray!15}{82.67}  & \cellcolor{gray!15}{95.33}& \cellcolor{gray!15}{78.00}\\
\hline
\multirow{6}{*}{\makecell[c]{Wooden level}}  
 &bowl            & 93.33 & 0.00 & 66.67 & 80.00 & 66.67   \\
 &boat            & 93.33 & 0.00 & 66.67 & 70.00 & 66.67  \\
 &chest           & 90.00 & 0.00 & 66.67 & 70.00 & 63.33\\
 &wooden sword    & 86.67 & 0.00 & 40.00 & 63.33 & 60.00       \\
 &wooden pickaxe  & 80.00 & 0.00 & 40.00 & 60.00 & 60.00     \\
 &\cellcolor{gray!15}{Average} & \cellcolor{gray!15}{88.67} & \cellcolor{gray!15}{0.00} & \cellcolor{gray!15}{56.00} & \cellcolor{gray!15}{68.67}& \cellcolor{gray!15}{63.33} \\
\hline
\multirow{6}{*}{\makecell[c]{Stone level}}  
 &cobblestone     & 80.00 & 0.00 & 10.00 & 50.00 & 60.00  \\
 &furnace         & 80.00 & 0.00 & 3.33 & 0.00 & 60.00  \\
 &stone pickaxe   & 80.00 & 0.00 & 0.00 & 0.00 & 56.67  \\
 &iron ore        & 60.00 & 0.00 & 0.00 & 0.00 & 40.00 \\
 &glass           & 80.00 & 0.00 & 0.00 & 0.00 & 43.33  \\
&\cellcolor{gray!15}{Average} & \cellcolor{gray!15}{76.00} & \cellcolor{gray!15}{0.00} & \cellcolor{gray!15}{2.67} & \cellcolor{gray!15}{10.00}& \cellcolor{gray!15}{52.00}  \\
\hline
\multirow{6}{*}{Iron level} 
 &iron ingot       & 56.67 & 0.00 & 0.00 & 0.00 & 36.67    \\
 &shield          & 56.67 & 0.00 & 0.00 & 0.00 & 36.67 \\
 &bucket          & 53.33 & 0.00 & 0.00 & 0.00 & 30.00   \\
 &iron pickaxe    & 50.00 & 0.00 & 0.00 & 0.00 & 26.67   \\
 &iron door       & 43.33 & 0.00 & 0.00 & 0.00 & 20.00 \\
 &\cellcolor{gray!15}{Average} & \cellcolor{gray!15}{52.00} & \cellcolor{gray!15}{0.00} & \cellcolor{gray!15}{0.00} & \cellcolor{gray!15}{0.00} & \cellcolor{gray!15}{30.00} \\
\hline
\multirow{6}{*}{Diamond level} 
 &diamond ore     & 30.00 & 0.00 & 0.00 & 0.00 & 10.00    \\
 &mind redstone   & 20.00 & 0.00 & 0.00 & 0.00 & 3.33    \\
 &compass         & 16.67 & 0.00 & 0.00 & 0.00 & 0.00    \\
 &diamond pickaxe & 23.33 & 0.00 & 0.00 & 0.00 & 3.33    \\
 &piston          & 20.00 & 0.00 & 0.00 & 0.00 & 3.33    \\
 &\cellcolor{gray!15}{Average} & \cellcolor{gray!15}{22.00} & \cellcolor{gray!15}{0.00} & \cellcolor{gray!15}{0.00} & \cellcolor{gray!15}{0.00}& \cellcolor{gray!15}{4.00}  \\
\bottomrule[1pt]

\end{tabular}
\label{tab:ablation for all components in sup}
\end{table*}

\begin{table*}[htbp]
\centering
\caption{More detailed success rates for different LLMs as zero-shot Planners on \textit{Process-Dependent Tasks}. The parts with a gray background in the table represent the average success rate for the current level.}
\begin{tabular}{cc|cccc}
\bottomrule[1pt]
\multirow{2}{*}{Task Level} & \multirow{2}{*}{Object} & \multicolumn{4}{c}{Success rate(\%)} \\

& & GPT-4(Ours) & GPT-3.5-Turbo~\cite{chatgpt} & LLaMA2-70B-Chat~\cite{touvron2023llama2} & Vicuna-13B-v1.5-16k~\cite{chiang2023vicuna}  \\
\hline
\multirow{6}{*}{\makecell[c]{Basic level}} 
 &log             & 96.67 & 96.67 & 6.67 & 3.33 \\
 &sand            & 96.67 & 96.67 & 3.33 & 3.33 \\
 &planks          & 96.67 & 96.67 & 0.00 & 0.00  \\
 &stick           & 96.67 & 96.67 & 0.00 & 0.00   \\
 &crafting table  & 93.33 & 90.00 & 0.00 & 0.00   \\
 &\cellcolor{gray!15}{Average} & \cellcolor{gray!15}{96.00} & \cellcolor{gray!15}{95.33} & \cellcolor{gray!15}{2.00} & \cellcolor{gray!15}{1.33} \\
\hline
\multirow{6}{*}{\makecell[c]{Wooden level}}  
 &bowl            & 93.33 & 90.00 & 0.00 & 0.00 \\
 &boat            & 93.33 & 90.00 & 0.00 & 0.00  \\
 &chest           & 90.00 & 90.00 & 0.00 & 0.00 \\
 &wooden sword    & 86.67 & 83.33 & 0.00 & 0.00 \\
 &wooden pickaxe  & 80.00 & 80.00 & 0.00 & 0.00  \\
 &\cellcolor{gray!15}{Average} & \cellcolor{gray!15}{88.67} & \cellcolor{gray!15}{86.67} & \cellcolor{gray!15}{0.00} & \cellcolor{gray!15}{0.00}\\
\hline
\multirow{6}{*}{\makecell[c]{Stone level}}  
 &cobblestone     & 80.00 & 66.67  & 0.00 & 0.00 \\
 &furnace         & 80.00 & 50.00  & 0.00 & 0.00   \\
 &stone pickaxe   & 80.00 & 50.00  & 0.00 & 0.00 \\
 &iron ore        & 60.00  & 10.00  & 0.00 & 0.00  \\
 &glass           & 80.00 & 33.33  & 0.00 & 0.00    \\
&\cellcolor{gray!15}{Average} & \cellcolor{gray!15}{76.00} & \cellcolor{gray!15}{42.00} & \cellcolor{gray!15}{0.00} & \cellcolor{gray!15}{0.00}\\
\hline
\multirow{6}{*}{Iron level} 
 &iron ingot       & 56.67 & 6.67 & 0.00 & 0.00   \\
 &shield          & 56.67 & 3.33 & 0.00 & 0.00 \\
 &bucket          & 53.33 & 0.00 & 0.00 & 0.00  \\
 &iron pickaxe    & 50.00 & 3.33 & 0.00 & 0.00   \\
 &iron door       & 43.33 & 0.00 & 0.00 & 0.00 \\
 &\cellcolor{gray!15}{Average} & \cellcolor{gray!15}{52.00} & \cellcolor{gray!15}{2.67} & \cellcolor{gray!15}{0.00} & \cellcolor{gray!15}{0.00} \\
\hline
\multirow{6}{*}{Diamond level} 
 &diamond ore     & 30.00 & 0.00 & 0.00 & 0.00  \\
 &mind redstone   & 20.00 & 0.00 & 0.00 & 0.00 \\
 &compass         & 16.67 & 0.00 & 0.00 & 0.00  \\
 &diamond pickaxe & 23.33 & 0.00 & 0.00 & 0.00   \\
 &piston          & 20.00 & 0.00 & 0.00 & 0.00       \\
 &\cellcolor{gray!15}{Average} & \cellcolor{gray!15}{22.00} & \cellcolor{gray!15}{0.00} & \cellcolor{gray!15}{0.00} & \cellcolor{gray!15}{0.00} \\
\bottomrule[1pt]

\end{tabular}
\label{tab:ablation for planner in sup}
\end{table*}

\begin{table*}[htbp]
\centering
\caption{Success rates for different parts of Memory on \textit{Process-Dependent Tasks}. The parts with a gray background in the table represent the average success rate for the current level.}
\begin{tabular}{cc|cccc}
\bottomrule[1pt]

\multirow{2}{*}{Task Level} & \multirow{2}{*}{Object} & \multicolumn{4}{c}{Success rate(\%)} \\

& & All Memory(Ours) &  w/o Performer Memory &  w/o Knowledge Memory &  w/o All Memory  \\
\hline
\multirow{6}{*}{\makecell[c]{Basic level}} 
 &log             & 96.67 & 96.67 & 90.00 & 90.00\\
 &sand            & 96.67 & 96.67 & 90.00 & 90.00\\
 &planks          & 96.67 & 96.67 & 83.33 & 80.00\\
 &stick           & 96.67 & 96.67 & 76.67 & 76.67   \\
 &crafting table  & 93.33 & 93.33 & 80.00 & 76.67   \\
 &\cellcolor{gray!15}{Average} & \cellcolor{gray!15}{96.00} & \cellcolor{gray!15}{96.00} & \cellcolor{gray!15}{84.00} & \cellcolor{gray!15}{82.67}\\
\hline
\multirow{6}{*}{\makecell[c]{Wooden level}}  
 &bowl            & 93.33 & 93.33 & 70.00 & 66.67\\
 &boat            & 93.33 & 90.00 & 66.67 & 66.67  \\
 &chest           & 90.00 & 90.00 & 70.00 & 66.67\\
 &wooden sword    & 86.67 & 83.33 & 43.33 & 40.00\\
 &wooden pickaxe  & 80.00 & 80.00 & 40.00 & 40.00\\
 &\cellcolor{gray!15}{Average} & \cellcolor{gray!15}{88.67} & \cellcolor{gray!15}{87.33} & \cellcolor{gray!15}{58.00} & \cellcolor{gray!15}{56.00}\\
\hline
\multirow{6}{*}{\makecell[c]{Stone level}}  
 &cobblestone     & 80.00 & 73.33  & 16.67 & 10.00 \\
 &furnace         & 80.00 & 73.33  & 6.67  & 3.33  \\
 &stone pickaxe   & 80.00 & 70.00  & 3.33  & 0.00 \\
 &iron ore        & 60.00 & 50.00  & 0.00  & 0.00 \\
 &glass           & 80.00 & 70.00  & 3.33  & 0.00  \\
&\cellcolor{gray!15}{Average} & \cellcolor{gray!15}{76.00} & \cellcolor{gray!15}{67.33} & \cellcolor{gray!15}{6.00} & \cellcolor{gray!15}{2.67}\\
\hline
\multirow{6}{*}{Iron level} 
 &iron ingot       & 56.67 & 53.33 & 0.00 & 0.00  \\
 &shield          & 56.67 & 53.33 & 0.00 & 0.00\\
 &bucket          & 53.33 & 46.67 & 0.00 & 0.00\\
 &iron pickaxe    & 50.00 & 43.33 & 0.00 & 0.00 \\
 &iron door       & 43.33 & 40.00 & 0.00 & 0.00 \\
 &\cellcolor{gray!15}{Average} & \cellcolor{gray!15}{52.00} & \cellcolor{gray!15}{47.33} & \cellcolor{gray!15}{0.00}  & \cellcolor{gray!15}{0.00}\\
\hline
\multirow{6}{*}{Diamond level} 
 &diamond ore     & 30.00 & 23.33 & 0.00 & 0.00  \\
 &mind redstone   & 20.00 & 26.67 & 0.00 & 0.00  \\
 &compass         & 16.67 & 10.00 & 0.00 & 0.00  \\
 &diamond pickaxe & 23.33 & 20.00 & 0.00 & 0.00  \\
 &piston          & 20.00 & 13.33 & 0.00 & 0.00     \\
 &\cellcolor{gray!15}{Average} & \cellcolor{gray!15}{22.00} & \cellcolor{gray!15}{16.67} & \cellcolor{gray!15}{0.00}  & \cellcolor{gray!15}{0.00}\\
\bottomrule[1pt]

\end{tabular}
\label{tab:ablation for memory in sup}
\end{table*}

\begin{table*}[htbp]
\centering
\caption{Success rates with and without the check part of the Patroller in the presence of \textit{``Random Drop''} Setting on \textit{Process-Dependent Tasks}. The parts with a gray background in the table represent the average success rate for the current level.}
\begin{tabular}{cc|cccc}
\bottomrule[1pt]

\multicolumn{2}{c|}{Component}  & \multicolumn{4}{c}{Method}  \\
\arrayrulecolor{black}\hline

\multicolumn{2}{c|}{the check part of Patroller}         & \checkmark & \ding{55} & \checkmark & \ding{55} \\
 
\multicolumn{2}{c|}{\textit{``Random Drop''}}            & \ding{55} & \ding{55} & \checkmark & \checkmark  \\

\bottomrule[1pt]

Task Level & Object & \multicolumn{4}{c}{ Success rate(\%)}  \\
\hline
\multirow{6}{*}{\makecell[c]{Basic level}} 
 &log             & 96.67 & 86.67 & 90.00 & 90.00     \\
 &sand            & 96.67 & 73.33 & 90.00 & 90.00  \\
 &planks          & 96.67 & 83.33 & 86.67 & 70.00    \\
 &stick           & 96.67 & 73.33 & 86.67 & 50.00  \\
 &crafting table  & 93.33 & 73.33 & 83.33 & 50.00   \\
 &\cellcolor{gray!15}{Average}& \cellcolor{gray!15}{96.00}& \cellcolor{gray!15}{78.00}& \cellcolor{gray!15}{78.00}& \cellcolor{gray!15}{70.00}\\
\hline
\multirow{6}{*}{\makecell[c]{Wooden level}}  
 &bowl            & 93.33 & 66.67 & 80.00 & 10.00 \\
 &boat            & 93.33 & 66.67 & 83.33 & 10.00\\
 &chest           & 90.00 & 63.33 & 80.00 & 10.00\\
 &wooden sword    & 86.67 & 60.00 & 70.00 & 3.33     \\
 &wooden pickaxe  & 80.00 & 60.00 & 70.00 & 3.33   \\
 &\cellcolor{gray!15}{Average}& \cellcolor{gray!15}{88.67} & \cellcolor{gray!15}{63.33}& \cellcolor{gray!15}{78.00}& \cellcolor{gray!15}{7.33} \\
\hline
\multirow{6}{*}{\makecell[c]{Stone level}}  
 &cobblestone     & 80.00 & 60.00  & 53.33 & 3.33\\
 &furnace         & 80.00 & 60.00  & 53.33 & 0.00\\
 &stone pickaxe   & 80.00 & 56.67  & 50.00 & 0.00\\
 &iron ore        & 60.00 & 40.00  & 30.00 & 0.00\\
 &glass           & 80.00 & 43.33  & 40.00 & 0.00\\
&\cellcolor{gray!15}{Average}& \cellcolor{gray!15}{76.00} & \cellcolor{gray!15}{52.00}& \cellcolor{gray!15}{45.33}& \cellcolor{gray!15}{0.00}  \\
\hline
\multirow{6}{*}{Iron level} 
 &iron ingot       & 56.67 & 36.67 & 26.67 & 0.00  \\
 &shield          & 56.67 & 36.67 & 26.67 & 0.00  \\
 &bucket          & 53.33 & 30.00 & 16.67 & 0.00  \\
 &iron pickaxe    & 50.00 & 26.67 & 13.33 & 0.00  \\
 &iron door       & 43.33 & 20.00 & 10.00 & 0.00  \\
 &\cellcolor{gray!15}{Average}& \cellcolor{gray!15}{52.00} & \cellcolor{gray!15}{30.00}& \cellcolor{gray!15}{18.67}& \cellcolor{gray!15}{0.00} \\
\hline
\multirow{6}{*}{Diamond level} 
 &diamond ore     & 30.00 & 10.00 & 3.33 & 0.00  \\
 &mind redstone   & 20.00 & 3.33  & 3.33 & 0.00 \\
 &compass         & 16.67 & 0.00  & 0.00 & 0.00 \\
 &diamond pickaxe & 23.33 & 3.33  & 0.00 & 0.00 \\
 &piston          & 20.00 & 3.33  & 0.00 & 0.00 \\
 &\cellcolor{gray!15}{Average}& \cellcolor{gray!15}{22.00} & \cellcolor{gray!15}{4.00}& \cellcolor{gray!15}{1.33}& \cellcolor{gray!15}{0.00}  \\
\bottomrule[1pt]

\end{tabular}
\label{tab:ablation for all patroller in sup}
\end{table*}

\end{document}